%% file: acl.tex
\pdfoutput=1
\documentclass[11pt]{article}

\usepackage[preprint]{acl}
\usepackage{times}
\usepackage{latexsym}

\usepackage[T1]{fontenc}

\usepackage[utf8]{inputenc}

\usepackage{microtype}

\usepackage{inconsolata}

\usepackage{graphicx}
\usepackage{amsmath}
\usepackage{amssymb}
\usepackage{pifont}
\usepackage{booktabs}
\usepackage{algorithm}
\usepackage{algpseudocode}
\algrenewcommand\algorithmicrequire{\textbf{Input:}}
\algrenewcommand\algorithmicensure{\textbf{Output:}}
\usepackage{algpseudocode}
\usepackage{threeparttable}
\usepackage{multirow}
\usepackage{tabularx}
\usepackage{wrapfig}
\usepackage[capitalize,noabbrev]{cleveref}
\usepackage{enumitem}
\usepackage[textsize=small,color=gray]{todonotes}
\usepackage{mathtools}
\usepackage{soul}
\usepackage{multirow}
\usepackage{colortbl}
\usepackage{tcolorbox}

\crefname{section}{\S}{\S\S}

\newcommand{\name}{IPR} 
\newcommand{\interalia}{\textit{inter alia}}

\title{IPR: Intelligent Prompt Routing with User-Controlled \\ Quality-Cost Trade-offs}

\author{
Aosong Feng\thanks{Equal contribution}, Balasubramaniam Srinivasan\footnotemark[1], Yun Zhou\footnotemark[1], Zhichao Xu\footnotemark[1], Kang Zhou, Sheng Guan,\\
{\bf Yueyan Chen}, {\bf Xian Wu}, {\bf Ninad Kulkarni}, {\bf Yi Zhang}, {\bf Zhengyuan Shen}, {\bf Dmitriy Bespalov}, \\
{\bf Soumya Smruti Mishra}, {\bf Yifei Teng}, {\bf Darren Yow-Bang Wang}, {\bf Haibo Ding\thanks{Corresponding author: hbding@amazon.com}}, {\bf Lin Lee Cheong} \\
AWS AI \\
\texttt{\{asfeng, srbalasu, yunzzhou, xzhichao, zhoukang, shguan, yyanc, xianwwu, ninadkul,}\\
\texttt{imyi, donshen, dbespal, soumish, yifeit, ybwang, hbding, lcheong\}@amazon.com }
}

\begin{document}
\maketitle
\begin{abstract}
Routing incoming queries to the most cost-effective LLM while maintaining response quality poses a fundamental challenge in optimizing performance-cost trade-offs for large-scale commercial systems.
We present IPR\,---\,a quality-constrained \textbf{I}ntelligent \textbf{P}rompt \textbf{R}outing framework that dynamically selects optimal models based on predicted response quality and user-specified tolerance levels.
IPR introduces three key innovations: (1) a modular architecture with lightweight quality estimators trained on 1.5M prompts annotated with calibrated quality scores, enabling fine-grained quality prediction across model families; (2) a user-controlled routing mechanism with tolerance parameter $\tau \in [0,1]$ that provides explicit control over quality-cost trade-offs; and (3) an extensible design using frozen encoders with model-specific adapters, reducing new model integration effort. 
To rigorously train and evaluate IPR, we detail our efforts to curate an IPR dataset containing 1.5 million examples with response quality annotations across 11 LLM candidates.
Deployed on a major cloud platform, IPR achieves 43.9\% cost reduction while maintaining quality parity with the strongest model in the Claude family and processes requests with consistently low latency.
The deployed system and additional product details are publicly available at \url{https://aws.amazon.com/bedrock/intelligent-prompt-routing/}.

\end{abstract}

\section{Introduction}
\label{sec:intro}

The proliferation of large language models (LLMs) with varying capabilities and costs has created a fundamental challenge in production deployments: how to automatically route incoming queries to the most cost-effective model while maintaining acceptable response quality \cite{hu2024routerbench}.
This challenge is exemplified in multi-model platforms like Amazon Bedrock, where models range from lightweight options like Claude Haiku to state-of-the-art models like Claude-3.5-Sonnet\,---\,a 12x cost difference. 
User queries exhibit enormous diversity: simple factual questions can be handled by smaller models, while complex reasoning tasks require more performant ones~\cite{jaech2024openaio1systemcard,guo2025deepseek}. 
However, existing systems either force users to manually select models, or employ rigid routing rules that fail to capture the continuous spectrum of query complexity, resulting in substantial unnecessary costs or degraded user experiences at scale~\cite{lu-etal-2024-routingtotheexpert,ding2024hybridllmcostefficientandqualityawarequeryrouting,ong2024routellm}.

\begin{figure}[t]
    \centering
    \includegraphics[width=0.95\linewidth]{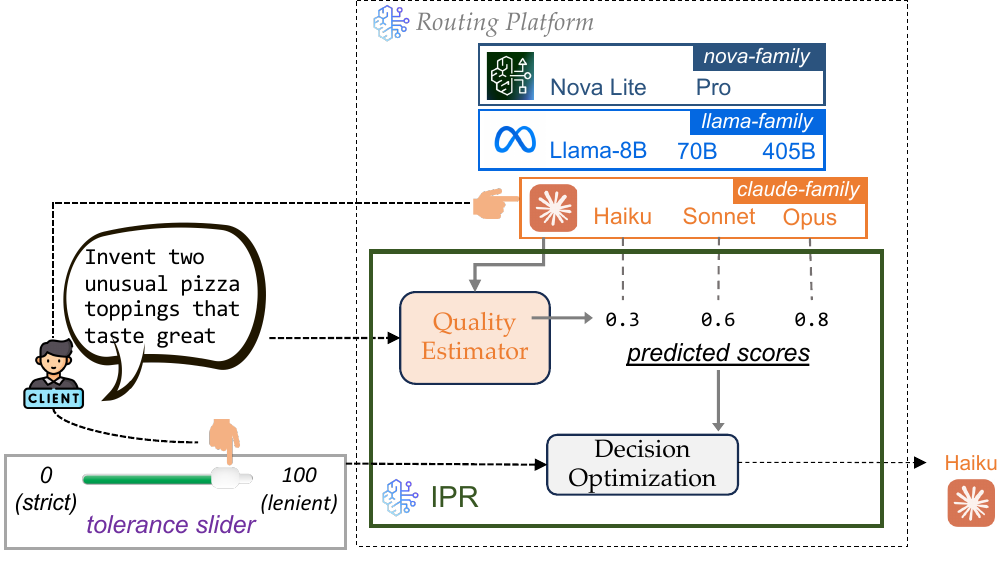}
    \caption{Routing prompt with \name~with user tolerance. }
    \label{fig:system_overview}\vspace{-0.3cm}
\end{figure}

Deploying prompt routing onto real-world production systems poses several fundamental challenges which existing approaches have not addressed comprehensively.
(1) \textbf{Quality prediction without response generation} requires routers to estimate response quality for each candidate model using only the input prompt, without actually generating responses\,---\,a challenging task given that model capabilities vary across different query types and domains.
While recent works like RouteLLM~\cite{ong2024routellm} explored lightweight BERT-based classifiers for this prediction, they either support only binary strong/weak decisions rather than continuous quality estimation or are trained on small-scale datasets.
(2) \textbf{Latency constraints} in production systems prevent use of approaches that require multiple model invocations or complex computations; cascade-based methods circumvent this by sequential evaluation but sacrifice flexibility in model selection \cite{yue2024largelanguagemodelcascades,chen2023frugalgpt}.
(3) \textbf{Model extensibility and diversity} becomes critical as production platforms must simultaneously support diverse model families (e.g., Nova, Claude, Llama) each with distinct characteristics, while seamlessly integrating frequent model updates and releases. 
Most existing routers require complete retraining when model portfolios change \cite{lu-etal-2024-routingtotheexpert,ding2024hybridllmcostefficientandqualityawarequeryrouting}, making them impractical for dynamic environments like a centralized LLM inference platform where new model versions appear monthly. 
(4) \textbf{User-specific quality-cost preferences} vary significantly across applications\,---\,a financial analysis task may prioritize accuracy while a chatbot may favor cost efficiency\,---\,yet current routing solutions offer little user control, typically hard-coding fixed quality thresholds \cite{ding2024hybridllmcostefficientandqualityawarequeryrouting} that cannot adapt to diverse business requirements.

To tackle these challenges, we propose \textbf{Intelligent Prompt Routing (IPR)}, a quality-constrained framework that dynamically selects the most cost-effective model while satisfying user-specified quality requirements. Our contributions are:

\begin{itemize}[leftmargin=*, topsep=0pt, itemsep=0pt]
\item \textbf{Industrial-scale quality prediction}: IPR trains neural estimators on 1.5M prompts annotated with calibrated reward scores from all candidate models, enabling fine-grained quality estimation that achieves 43.9\% cost reduction in Claude family while maintaining quality parity.
\item \textbf{Efficient extensible architecture}: A modular design with backbone prompt encoder and lightweight candidate-specific adapters enables fast routing decisions. 
\item \textbf{User-controlled routing}: A quality tolerance parameter ($\tau \in [0,1]$) provides explicit control over cost-quality trade-offs, from maximum quality ($\tau=0$) to aggressive savings ($\tau=1$), with dynamic per-prompt threshold adjustment.
\end{itemize}

\begin{figure*}[btp]
    \centering
    \includegraphics[width=\linewidth]{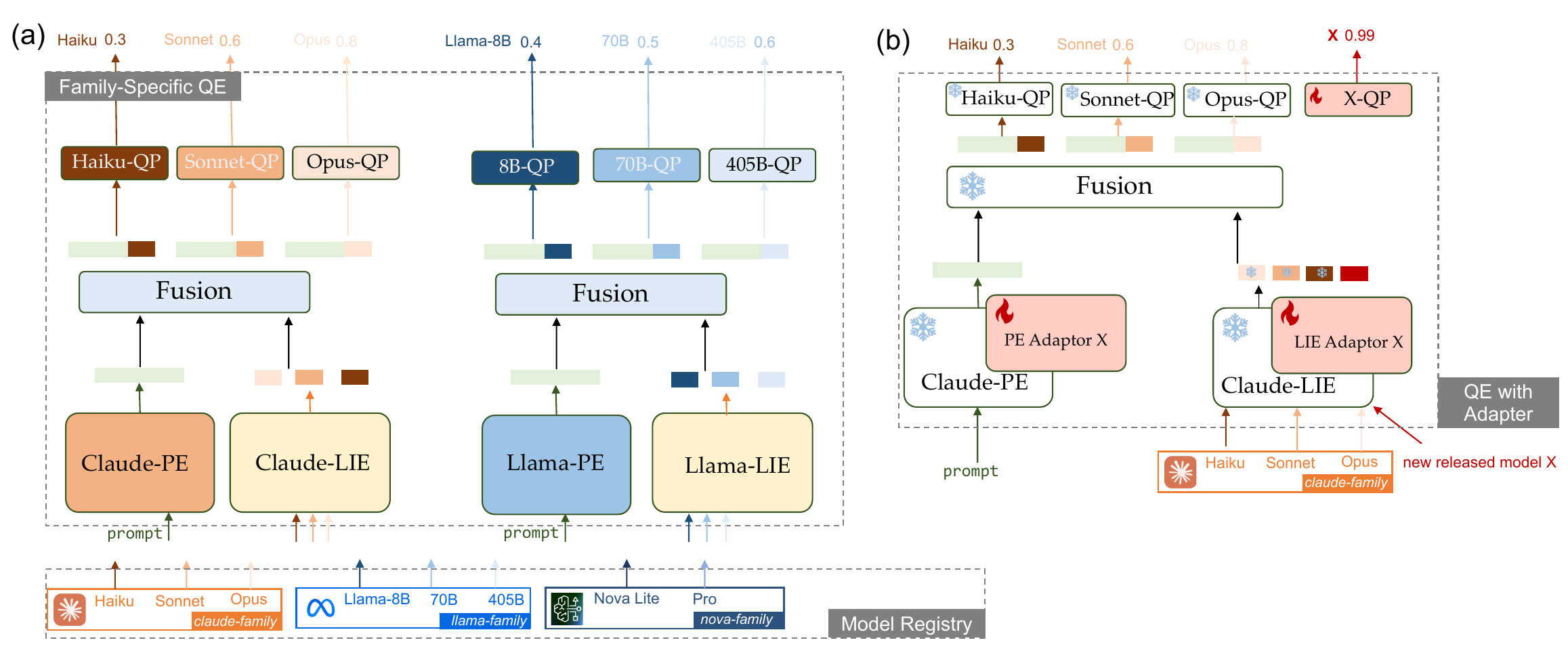}
    \caption{IPR Quality Estimator (QE) architecture. (a) Family-specific Quality Predictors (QP) process prompts with candidate model IDs to predict quality scores. (b) New models are integrated via lightweight adapters on frozen Prompt Encoder (PE) and LLM Identity Encoder (LIE) components.}
    \vspace{-8pt}
    \label{fig:ipr_model_architecture}
\end{figure*}

\section{Problem Formulation and Routing Framework}
\label{sec:problem_formulation}
We present a formal treatment of the prompt routing problem focusing on performance-efficiency trade-offs. Given a user prompt and candidate LLMs, we aim to select the most cost-efficient model whose predicted response quality satisfies user-specified response quality tolerance.

\subsection{LLM Routing Formulation}
\label{subsec:problem_formulation}
We formulate LLM routing as a constrained optimization problem. Denote the set of candidate LLMs as $\mathcal{C}$. Given a prompt $x_i$, each candidate model $c \in \mathcal{C}$ would generate a response $y_{i,c} = c(x_i)$ with quality $r_{i,c} = R(x_i, y_{i,c}) \in [0, 1]$ measured by a reward function capturing human preference alignment. The cost of invoking model $c$ is denoted by $v_{i,c}$.

To make the quality-cost trade-off explicit and controllable, we minimize cost subject to quality constraints. Users specify a quality tolerance $\tau \in [0, 1]$ defining acceptable quality degradation relative to the best available model. This induces a feasible set: 
$\mathcal{C}_\tau = \left\{ c \in \mathcal{C} \;\middle|\; G(r_{i,c}, \tau) \geq 0 \right\}$,
where $G$ is a performance-gating function determining whether candidate performance satisfies user tolerance. The optimal routing decision selects the most efficient model from the feasible set:
\begin{equation}
\label{eq:argmin_cost}
c^*_i = \arg\min_{c \in \mathcal{C}_\tau} v_{i,c}.
\end{equation}
Since computing $r_{i,c}$ requires generating and evaluating responses, we train a quality estimator $\hat{r}_{i,c} = R_\theta(x_i, c)$ that predicts response quality using only the prompt and candidate identity:
\begin{equation}
\theta^* = \arg\min_\theta \sum_{i,c} \ell(R_\theta(x_i, c), r_{i,c}),
\end{equation}
where $\ell$ is a regression loss and $r_{i,c}$ are ground-truth rewards from a calibrated reward model.

\subsection{Routing Strategy}
\label{subsec:routing}
Given the quality estimates $\hat{r}_{i,c}$, we implement a quality-constrained, cost-optimal routing strategy through two stages: (1) filter out candidates not meeting quality tolerance, (2) select the most cost-efficient qualified model.

The gating function translates user tolerance $\tau$ to a quality threshold:
\begin{align}
G(\hat{r}_{i,c}, \tau) &= \hat{r}_{i,c} - r_{i,\text{th}} \geq 0, \\
\text{where} \quad r_{i,\text{th}} &= \hat{r}_{i,\text{max}} - \tau \cdot (\hat{r}_{i,\text{max}} - \hat{r}_{i,\text{min}}).
\end{align}
Here, $\hat{r}_{i,\text{min}}$ is designed for min-max scaling when predictions have a non-zero baseline.
We use dynamic per-prompt $\hat{r}_{i,\text{max}} = \max_{c \in \mathcal{C}} \hat{r}_{i,c}$ to handle varying query complexity and fixed $\hat{r}_{i,\text{min}} = 0$ for stability. This hybrid strategy adapts to individual prompt difficulty while preventing threshold collapse (More routing strategies are discussed in Appendix \ref{asec:ablation_studies}).
Once the feasible set $\mathcal{C}_\tau(x_i)$ is determined, we select the minimum-cost candidate following~\cref{eq:argmin_cost}.

\subsection{Evaluation}
\label{subsec:evaluation}
We evaluate IPR along two dimensions: quality prediction accuracy and end-to-end routing performance. For quality prediction, we use mean absolute error and ranking-based metrics, including Top-K accuracy and F1 scores. For routing performance, we introduce a area-under-the-curve (AUC) style metric named Bounded-ARQGC (and its variant, Relative-ARQGC) to measure quality-cost trade-offs across all tolerance settings and Cost Save Ratio (CSR) for different settings. Detailed metric definitions are provided in~\cref{asec:evaluation_metrics}.


\begin{algorithm}[h]
\caption{IPR Routing with User Tolerance}
\label{alg:ipr}
\begin{algorithmic}[1]
\Require Prompt $x$, candidate set $\mathcal{C}$, prices $\{v_c\}$, tolerance $\tau\in[0,1]$, QE $R_\theta$, safety margin $\delta\ge 0$
\Ensure Routed model $c^\star$
\State $\mathbf{p} \leftarrow \mathrm{PE}(x)$ \Comment{Prompt embedding (cached across turns if multi‑turn)}
\For{$c \in \mathcal{C}$}
  \State $\mathbf{e}_c \leftarrow \mathrm{LIE}(c)$ \Comment{Model identity embedding}
  \State $\hat r_c \leftarrow \mathrm{QP}([\mathbf{p}; \mathbf{e}_c])$ \Comment{Predicted quality (optionally calibrated)}
\EndFor
\State $\hat r_{\max} \leftarrow \max_{c\in\mathcal{C}} \hat r_c$
\State $r_{\mathrm{th}} \leftarrow (1-\tau)\cdot \hat r_{\max}$ \Comment{Per‑prompt threshold with safety margin}
\State $\mathcal{F} \leftarrow \{ c \in \mathcal{C} \mid \hat r_c \ge r_{\mathrm{th}}\}$ \Comment{Feasible candidates}
\If{$\mathcal{F}=\emptyset$}
  \State $\mathcal{F} \leftarrow \{\arg\max_{c} \hat r_c\}$ \Comment{Fallback to predicted best}
\EndIf
\State $c^\star \leftarrow \arg\min_{c \in \mathcal{F}} v_c$ \Comment{Minimize monetary cost; tie‑break by $\hat r_c$}
\State \Return $c^\star$
\end{algorithmic}
\end{algorithm}

\section{Intelligent Prompt Routing}
\label{sec:methodology}

\subsection{System Overview}
\label{subsec:system_overview}

Figure \ref{fig:system_overview} illustrates the IPR platform, comprising three core components: (1) the \textit{Quality Estimator} (QE) that predicts response quality for each candidate model, (2) the \textit{Decision Optimization} (DO) module that executes quality-constrained routing decisions, and (3) the \textit{Model Registry} that maintains model metadata and configurations.

The routing pipeline operates as follows: Upon receiving a user prompt with optional multi-turn context, the system captures the user's quality-cost preference through tolerance parameter $\tau \in [0, 1]$ (where $\tau = 0$ enforces maximum quality and $\tau = 1$ maximizes cost savings). The Quality Estimator computes predicted quality scores $\hat{r}_{i,c}$ for each candidate $c \in \mathcal{C}$ using the learned estimator $R_\theta(x_i, c)$. These predictions feed into the Decision Optimization module, which applies tolerance-based filtering to identify feasible candidates and selects the minimum-cost model.

Our architecture achieves two critical objectives: (1) \textit{extensibility} through lightweight adapters for new models without full retraining, and (2) \textit{efficiency} with fast routing decisions based on prompt embeddings. A sketch of the algorithm is shown in Algorithm~\ref{alg:ipr}.

\subsection{Quality Estimator Architecture}
\label{subsec:quality_estimator}

The Quality Estimator predicts scalar reward scores approximating response quality for each prompt-model pair. As shown in~\cref{fig:ipr_model_architecture}, it consists of three key components:

\textbf{(1) Prompt Encoder:} Maps input prompts to dense embeddings $\mathbf{p}_i = \mathrm{PE}(x_i) \in \mathbb{R}^d$ capturing semantic features relevant for quality prediction. We employ family-specific encoders (e.g., Claude-PE, Llama-PE) to capture model-specific patterns.

\textbf{(2) LLM Identity Encoder:} Provides learnable embeddings $\mathbf{e}_c = \mathrm{LIE}(c) \in \mathbb{R}^{d'}$ for each candidate model, encoding behavioral properties like verbosity and style.

\textbf{(3) Quality Predictor:} Fuses prompt and LLM embeddings via concatenation and predicts quality through a feed-forward network: $\hat{r}_{i,c} = \mathrm{QP}(\mathrm{Concat}(\mathbf{p}_i, \mathbf{e}_c))$.

We adopt family-specific architectures with separate prediction heads per model, enabling better within-family generalization and simplified integration of new models. Training uses \textbf{reward model scores} as supervision signals, providing fine-grained quality labels at scale (detailed in~\cref{asec:reward_modeling}).

IPR's modular design enables seamless integration of new LLMs without full retraining. When adding a new model, we freeze the core encoders and attach lightweight adapters that specialize the shared representations. This approach reduces integration and deployment time while preserving performance on existing models. Implementation details are provided in~\cref{asec:modular_adaptation}.

\section{Experiments}
\label{sec:experiment}
We train and evaluate IPR model on our dataset containing 1.5M prompts with quality annotations across multiple LLM families (details in~\cref{asec:dataset_collection}). This industrial-scale benchmark enables rigorous evaluation of routing systems under realistic conditions.

\input{tables/dataset_stat}

\subsection{Dataset Collection}
We construct the training and evaluation datasets using a diverse set of resources, covering open-domain dialogue, instruction-tuning, summarization, reasoning, and domain-specific question answering. 
The primary dataset includes responses from multiple language model families (Claude, Llama, Nova), where each instance contains outputs from all models within the same family, enabling direct comparison of response quality. 
Each response is annotated with a reward score assigned by the \texttt{Skywork/Skywork-Reward-Gemma-2-27B} model\footnote{\url{https://huggingface.co/Skywork/Skywork-Reward-Gemma-2-27B-v0.2}}~\cite{liu2024skyworkreward}, which serves as the supervision signal for training the quality estimator. The scale and split of the Combined dataset across different model families are summarized in~\cref{tab:dataset-sizes}. More details are listed in~\cref{asec:dataset_collection}.

\input{tables/tab-quality-estimation}
\input{figs/figure-quality-cost-curve}

\subsection{Experimental Setup}
\label{subsec:setup}

\paragraph{Model Families and Candidates.}
We evaluate IPR on three major LLM families, encompassing diverse model sizes and capabilities:
\begin{itemize}[leftmargin=*, topsep=0pt, itemsep=0pt]
\item \textbf{Claude family}: Claude-3-Haiku, Claude-3.5-Haiku, Claude-3.5-Sonnet variants
\item \textbf{Llama family}: Llama-3.1-\{8B, 70B\}, Llama-3.2-\{11B, 90B\}, Llama-3.3-70B
\item \textbf{Nova family}: Nova-Lite and Nova-Pro
\end{itemize}
These candidates represent different quality-cost trade-offs, allowing comprehensive routing evaluation. We chose to deploy family-specific routers due to their superior in-domain empirical performance (comparison of family-specific and unified router is shown in \cref{asec:ablation_studies}).

\input{tables/tab-overall-routing}
\input{tables/tab-routing-strategy}

\paragraph{Baseline Methods.}
We compare against: (1) \textbf{Static} routing to fixed models providing cost bounds, (2) \textbf{Random} uniform assignment, (3) \textbf{Oracle} routing with ground-truth quality scores, (4) \textbf{Budget-Aware Random} maintaining IPR's routing proportions but random assignment, and (5) \textbf{Classifier} following RouteLLM's approach.

\paragraph{Training Configuration.}
Models are trained on IPR train set (1.5M samples) using 8 or 16 A100 GPUs. 

\subsection{Results}
We evaluate \name~across three key dimensions: (1) quality estimation results, (2) overall routing performance across the full tolerance spectrum, and (3) cost savings at critical operating points. Our results demonstrate that for Claude family, \name~is able to achieve up to 43.9\% cost reduction for Stella embedding model when maintaining quality equivalent to the most expensive model, while providing flexible quality-cost trade-offs for diverse user preferences. 

\paragraph{Quality Estimation Performance.}
\cref{tab:quality_estimation_main} presents quality estimation results on IPR test set across different backbone architectures.
Scaling backbone size consistently improves prediction: Qwen3-emb-4B achieves lowest MAE (0.084 for Claude), 13.3\% better than RoBERTa-355M.
Embedding-based encoders outperform decoder counterparts at equivalent sizes. Notably, our selected Stella-400M achieves >73\% top-1 accuracy while being 8× faster than 4B models.

\paragraph{End-to-End Routing Performance.}
Table \ref{tab:routing_performance_main} presents the overall routing performance across the full tolerance spectrum, measured by our primary metric Bounded-ARQGC.
Across all model families, \name~variants significantly outperform baseline approaches, with the best configurations achieving 0.821 (Claude), 0.685 (Llama), and 0.766 (Nova) Bounded-ARQGC scores\,---\,representing relative improvements of 58.8\%, 38.1\%, and 57.6\% over random routing respectively.
The oracle router, which has access to ground-truth quality scores, establishes upper bounds of 0.915, 0.868, and 0.905, indicating room for future improvements in quality estimation.

Several key patterns emerge from these results.
First, modest quality estimation improvements yield disproportionate routing gains: Stella-400M's 2-7\% MAE reduction over RoBERTa translates to 3-21\% higher Bounded-ARQGC. Figure \ref{fig:quality_cost_curve_backbone} visualizes quality-cost trade-offs, showing IPR produces Pareto-optimal curves compared to baselines.
Second, the relationship between model scale and routing effectiveness exhibits diminishing returns\,---\,Qwen3-emb-4B improves MAE by 8-11\% over Stella-400M but yields only 2-16\% better routing performance, suggesting that accurate relative quality rankings matter more than precise score predictions.
Figure \ref{fig:quality_cost_curve_backbone} visualizes quality-cost trade-offs, showing IPR produces Pareto-optimal curves dominating baseline approaches.

\paragraph{Routing Latency and Efficiency.}
IPR’s routing decision requires a \emph{single} forward pass of the prompt encoder to compute a prompt embedding, followed by tiny per-candidate MLP heads; \emph{no autoregressive decoding} is involved. As a result, routing latency is \emph{input-length dependent} but \emph{output-length invariant}, and it adds only a few milliseconds before the selected endpoint is invoked.
To make efficiency concrete, we benchmark on \textbf{1$\times$ A100-40GB (PCIe), CUDA~12.4} with batch$=$1, FP32, 100 warmup steps and 1{,}000 measured runs per setting. We vary input length (500 to 1000 tokens) and candidate set size ($|\mathcal{C}|=5$ to  $10$). We report end-to-end wall-clock \textbf{P90/P99} (tokenization $\to$ encoder $\to$ heads $\to$ selection) and \textbf{peak memory} as in Table \ref{tab:router_latency}.

\begin{table}[t]
\centering
\caption{Router latency and memory (end-to-end, batch$=$1) measured on a single A100-40GB GPU. Latency calculation includes tokenization, encoder forward, per-candidate heads, and selection.}
\label{tab:router_latency}

\resizebox{\linewidth}{!}{%
\begin{tabular}{lcccccc}
\toprule
\textbf{Name}  & \textbf{Input (tok)} & $\boldsymbol{|\mathcal{C}|}$ & \textbf{P90 (ms)} & \textbf{P99 (ms)} & \textbf{Mem (GB)} \\
\midrule
IPR (Stella)         & 500  & 5  & 35.66  & 36.31  & 1.68 \\
IPR (Stella)         & 1000 & 5  & 64.92  & 65.14  & 1.72 \\
IPR (Stella)         & 1000 & 10 & 67.03  & 67.13  & 1.76 \\
IPR (Qwen3-0.6B)     & 500  & 5  & 56.79  & 62.69  & 3.02 \\
IPR (Qwen3-0.6B)     & 1000 & 5  & 115.30 & 116.74 & 3.80 \\
IPR (Qwen3-0.6B)     & 1000 & 10 & 118.54 & 119.02 & 3.83 \\
IPR (Qwen3-4B)       & 500  & 5  & 277.66 & 278.07 & 16.00 \\
IPR (Qwen3-4B)       & 1000 & 5  & 557.05 & 557.42 & 16.81 \\
IPR (Qwen3-4B)       & 1000 & 10 & 560.24 & 560.52 & 16.82 \\
\bottomrule
\end{tabular}
}

\end{table}

\paragraph{Performance at Critical Operating Points.}
While aggregate metrics capture overall routing effectiveness, practical deployment often focuses on specific quality-cost targets. Table \ref{tab:claude_iprbench_compact} examines router performance at two critical operating points: maintaining 100\% quality parity with the strongest model and accepting 5\% quality degradation. 
At the 100\% quality threshold\,---\,where users demand performance equivalent to always using the most capable model\,---\,IPR with Qwen3-0.6B achieves 48.7\% cost savings by routing 59.9\% of prompts to the more efficient Haiku model. This demonstrates that nearly 60\% of real-world prompts do not require the most expensive model to achieve optimal quality.
The routing distribution reveals how different backbones assess prompt complexity. Smaller encoders (RoBERTa-355M) exhibit more conservative routing, sending only 48.8\% to Haiku, while mid-sized models like Qwen3-0.6B achieve better prompt discrimination.

Results on IPR test dataset suggest IPR's effectiveness: achieving substantial cost reductions while maintaining quality, with flexible user control over trade-offs. Comprehensive ablation studies validating our design choices are provided in~\cref{asec:ablation_studies}.
To verify routing quality, we conduct blind human annotation studies detailed in Appendix \ref{asec:human_annotation}.

\section{Related Works}
\label{sec:related}
Here, we focus on existing prompt routing approaches, and defer benchmarks and evaluations to~\cref{asec:relatedworks}.

In the literature, different model designs and corresponding training strategies have been proposed for the LLM routing problem~\cite[\interalia]{lu-etal-2024-routingtotheexpert,ding2024hybridllmcostefficientandqualityawarequeryrouting,ong2024routellm,sikeridis2024pickllm,jitkrittum2025universalmodelroutingforefficientllminference,feng2025graphrouter,su2025cprouter,chuang2025learningtoroutellmswithconfidencetokens,stripelis2024tensoropera,mei2025omnirouter,sakota2024forc,chen2023frugalgpt,chen2024routerdc,jin2025radialrouter,ding2025bestroute,sikeridis2025pickllm,jitkrittum2025universalmodelroutingforefficientllminference,pan2025routetoreason,zhuang2024embedllm}.
HybridLLM~\cite{ding2024hybridllmcostefficientandqualityawarequeryrouting} employs a BERT-based encoder to optimize the cost-quality trade-off by routing "easy" queries to resource-efficient smaller models and "hard" queries to larger, more capable models.
EmbedLLM~\cite{zhuang2024embedllm} introduces a specialized encoder-decoder network for embedding LLM representations.
RouteLLM~\cite{ong2024routellm} implements a dynamic routing mechanism that intelligently routes prompts between a stronger and weaker LLM through various methodologies: similarity-weighted ranking,  matrix factorization, BERT-based classification, and Causal LLM classification. 
Zooter~\cite{lu-etal-2024-routingtotheexpert} also deploys reward model scores as the supervision signals and train the router with RankNet loss~\cite{burges2010ranknet}. Additionally, it leverages a tag-based label enhancement strategy to remove reward model noises.
GraphRouter~\cite{feng2025graphrouter} formulates LLM selection as edge prediction problem in a graph based framework and fully utilizes the information in the training data by jointly modeling the query-model, query-query, and model-model relationship. 
OmniRouter~\cite{mei2025omnirouter} formulates the routing task as a constrained optimization problem and leverages a hybrid retrieval-augmented predictor to predict the capabilities and costs of LLMs.
Different from aforementioned works that deploy clustering or train with teacher forcing, PickLLM~\cite{sikeridis2024pickllm} proposes a reinforcement learning-based routing framework that optimizes a composite reward function incorporating latency, computational cost, and response quality. 
IPR deploys a conventional supervised learning approach and focus on scaling training data mixture for robust LLM routing.

\section{Conclusions and Future Works}

We introduce Intelligent Prompt Routing\,---\,a low latency solution to cost efficient prompt routing. We detail our scientific experimentation: including curation of a large-scale training and evaluation dataset, design of evaluation metrics Bounded-ARQGC and CSR, different model architecture, training strategy ablations throughout the product development. 
Our future works will include incorporating multifaceted evaluations and supporting new model releases on our platform.

\section*{Limitations}
While IPR demonstrates strong performance in production deployment, several limitations merit discussion.
First, our quality estimation relies on reward model scores as supervision signals, which may not perfectly capture all aspects of human preference, particularly for specialized domains or creative tasks. 
Second, the current framework assumes independent routing decisions per prompt without considering conversation-level context or user session patterns, potentially missing optimization opportunities in multi-turn interactions. 
Third, our evaluation focuses on three model families (Claude, Llama, Nova) on a single platform; generalization to other model families or deployment environments requires further validation.
Finally, the modular adaptation mechanism, while efficient, still requires access to labeled data for new models, which may not be immediately available upon model release. Addressing these limitations, particularly through online learning from user feedback and session-aware routing, represents important directions for future work.

\bibliography{custom,anthology}

\appendix

\input{sec-appendix}

\end{document}

%% file: tables/dataset_stat.tex
\begin{table}[t]
\centering
\caption{IPR dataset size by model family and split. More details in Appendix~\ref{asec:dataset_collection}.}
\label{tab:dataset-sizes}
\resizebox{\linewidth}{!}{%
\begin{tabular}{llrrr}
\toprule
\textbf{Dataset} & \textbf{Subset} & \textbf{Claude} & \textbf{Llama} & \textbf{Nova} \\
\midrule
\multirow{3}{*}{Combined} & Training & 1,510,415 & 1,325,628 & 1,510,250 \\
                          & Dev      & 5,641     & 4,976     & 5,640     \\
                          & Test     & 5,642     & 5,032     & 5,641     \\
\midrule
MS Marco     & Test & 2,000 & 2,000 & 1,997 \\
Nvidia Chat  & Test & 2,000 & 2,000 & 1,999 \\
\bottomrule
\end{tabular}%
}

\end{table}

%% file: tables/tab-quality-estimation.tex
\begin{table*}[ht]
\label{tab:qe_results}
\centering
\caption{Quality estimation performance on IPR test set. We report Mean Absolute Error (MAE), Top-1 Accuracy, and F1 scores for different router architectures. Best results are \textbf{bolded}, second best are \underline{underlined}.}
\label{tab:quality_estimation_main}
\resizebox{\textwidth}{!}{%
\begin{tabular}{l|ccc|ccc|ccc}
\toprule
\multirow{2}{*}{\textbf{Method}} & \multicolumn{3}{c|}{\textbf{Claude}} & \multicolumn{3}{c|}{\textbf{Llama}} & \multicolumn{3}{c}{\textbf{Nova}} \\
& MAE $\downarrow$ & Top-1 $\uparrow$ & F1-macro $\uparrow$ & MAE $\downarrow$ & Top-1 $\uparrow$ & F1-macro $\uparrow$ & MAE $\downarrow$ & Top-1 $\uparrow$ & F1-macro $\uparrow$ \\
\midrule
IPR (RoBERTa-355M) & 0.09503 & 0.7025 & 0.6612 & 0.09283 & 0.7025 & 0.4825 & 0.09681 & 0.6500 & 0.5311\\
\rowcolor{gray!20}
IPR (Stella-400M) & 0.09478 & 0.7321 & 0.6629 & 0.08626 & 0.7154 & \underline{0.5139} & 0.09597 & 0.6408 & 0.5828 \\
\rowcolor{gray!20}
IPR (Qwen3-0.6B) & 0.09027 & 0.7353 & \underline{0.6934} & 0.09120 & \textbf{0.7257} & 0.4940 & 0.09509 & 0.6642 & 0.5834 \\
IPR (Qwen3-4B) & \underline{0.08540} & \underline{0.7463} & 0.6857 & \underline{0.08091} & 0.7237 & 0.5111 & \underline{0.08384} & \textbf{0.6853} & \underline{0.6016} \\
\rowcolor{gray!20}
IPR (Qwen3-emb-0.6B) & 0.08988 & 0.7408 & \textbf{0.6982} & 0.08963 & \underline{0.7217} & 0.4991 & 0.09279 & 0.6674 & 0.5717 \\
\rowcolor{gray!20}
IPR (Qwen3-emb-4B) & \textbf{0.08390} & \textbf{0.7508} & 0.6931 & \textbf{0.07997} & 0.7094 & \textbf{0.5664} & \textbf{0.08281} & \underline{0.6826} & \textbf{0.6070} \\
\bottomrule
\end{tabular}%
}

\end{table*}

%% file: figs/figure-quality-cost-curve.tex
\begin{figure*}[t]
    \centering
    \includegraphics[width=0.95\linewidth]{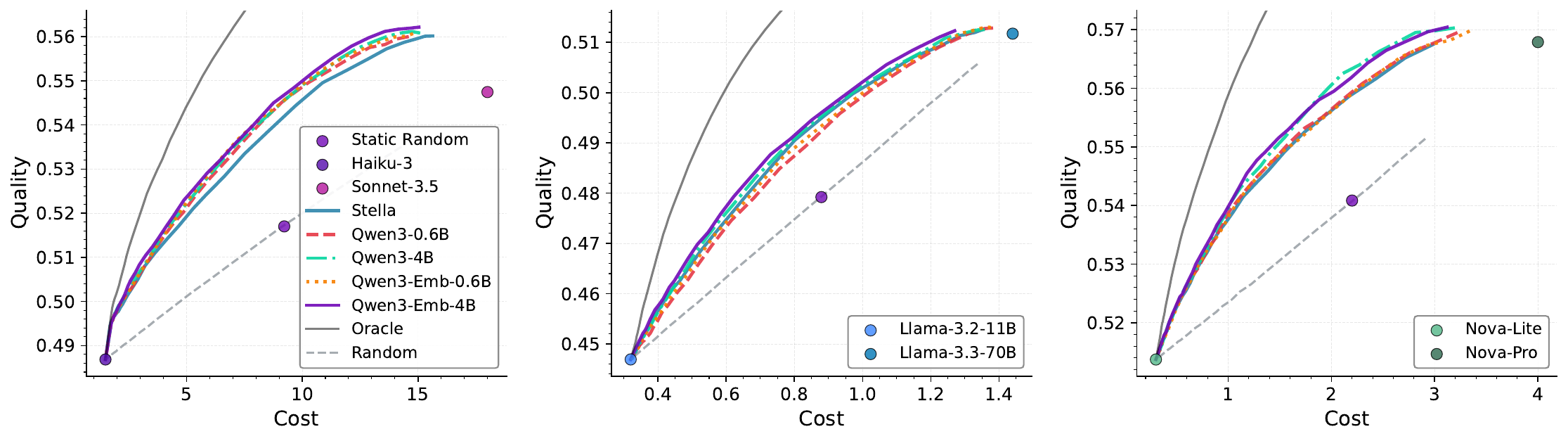}
    \caption{Quality and cost trade-offs under different user tolerance.}
    \label{fig:quality_cost_curve_backbone}
\end{figure*}

%% file: tables/tab-overall-routing.tex
\begin{table*}[ht]
\centering
\caption{Overall routing performance on IPR test set. We report Bounded-ARQGC (primary metric), Relative ARQGC improvement over random baseline. Best results (excluding Oracle) are \textbf{bolded}, second-best are \underline{underlined}. Rows with gray shading indicate encoder-based architectures.}
\label{tab:routing_performance_main}
\resizebox{\textwidth}{!}{%
\begin{tabular}{l|cc|cc|cc}
\toprule
\multirow{2}{*}{\textbf{Method}} & \multicolumn{2}{c|}{\textbf{Claude}} & \multicolumn{2}{c|}{\textbf{Llama}} & \multicolumn{2}{c}{\textbf{Nova}} \\
\cmidrule(lr){2-3} \cmidrule(lr){4-5} \cmidrule(lr){6-7}
& B-ARQGC $\uparrow$ & Rel-ARQGC $\uparrow$ & B-ARQGC $\uparrow$ & Rel-ARQGC $\uparrow$ & B-ARQGC $\uparrow$ & Rel-ARQGC $\uparrow$  \\
\midrule
\rowcolor{gray!10}
Oracle & 0.915 & 1.000 & 0.868 & 1.000 & 0.905 & 1.000  \\
\midrule
Static (Strongest) & - & -  & - & -  & - & -  \\
Static (Weakest) & - & -  & - & -  & - & -  \\
Random & 0.517 & 0.434 & 0.496 & 0.504  & 0.486 & 0.431  \\
\rowcolor{gray!10}
RouteLLM & 0.728 & 0.683 & 0.635 & 0.630 & 0.695 & 0.618 \\
\midrule
\rowcolor{gray!10}
IPR (Roberta-355M) & 0.732 & 0.695 & 0.628 & 0.625 & 0.707 & 0.622 \\
\rowcolor{gray!10}
IPR (Stella-400M) & 0.799 & 0.724 & 0.663 & 0.676 & 0.731 & 0.650 \\
IPR (Qwen3-0.6B) & 0.808 & 0.730 & 0.641 & 0.653 & 0.739 & 0.658 \\
IPR (Qwen3-4B) & 0.813 & \underline{0.743} & \underline{0.672} & \underline{0.685} & \textbf{0.766} & \textbf{0.687} \\
\rowcolor{gray!10}
IPR (Qwen3-emb-0.6B) & \underline{0.814} & 0.740 & 0.653 & 0.666 & 0.735 & 0.656 \\
\rowcolor{gray!10}
IPR (Qwen3-emb-4B) & \textbf{0.821} & \textbf{0.756} & \textbf{0.685} & \textbf{0.698} & \textbf{0.766} & \textbf{0.687}  \\
\bottomrule
\end{tabular}
}

\end{table*}

%% file: tables/tab-routing-strategy.tex
\begin{table*}[ht]
\centering
\caption{Router performance on IPR test dataset at 100\% and 95\% of the strongest model quality in Claude family.}
\label{tab:claude_iprbench_compact}
\resizebox{0.85\textwidth}{!}{
\begin{tabular}{l|cc|cc|cc|cc}
\toprule
\multirow{3}{*}{\textbf{Method}} & \multicolumn{4}{c|}{\textbf{100\% Quality}} & \multicolumn{4}{c}{\textbf{95\% Quality}} \\
\cmidrule(lr){2-5} \cmidrule(lr){6-9}
& \multirow{2}{*}{\textbf{CSR}} & \multirow{2}{*}{\textbf{Acc.}} & \multicolumn{2}{c|}{\textbf{Route Percentage}} & \multirow{2}{*}{\textbf{CSR}} & \multirow{2}{*}{\textbf{Acc.}} & \multicolumn{2}{c}{\textbf{Route Percentage}} \\
\cmidrule(lr){4-5} \cmidrule(lr){8-9}
& & & \textbf{Haiku} & \textbf{Sonnet} & & & \textbf{Haiku} & \textbf{Sonnet} \\
\midrule
\rowcolor{gray!10}
oracle & 0.705 & 1.0 & 60.43 & 39.57 & 0.685 & 1.0 & 77.27 & 22.72 \\
\rowcolor{gray!10}
RouteLLM & 0.425 &0.605 & 50.28 & 49.72 & 0.712 & 0.732 & 75.32 & 24.68 \\
\midrule
\rowcolor{gray!10}
IPR(RoBERTa) & 0.385 & 0.638 & 48.8 & 51.2 & 0.658 & 0.756 & 79.2 & 10.8\\
\rowcolor{gray!10}
IPR(Stella) & 0.439 & 0.678 & 54.41 & 45.59 & 0.730 & 0.811 & 82.69 & 17.30 \\
IPR(Qwen3-0.6B) & \textbf{0.487} & 0.688 & \textbf{59.95} & 40.04 & 0.730 & 0.799 & 83.69 & 16.30 \\
IPR(Qwen3-4B) & \underline{0.484} & \textbf{0.702} & \underline{57.95} & 42.04 & \underline{0.748} & \textbf{0.845} & \underline{84.01} & 15.99 \\
\rowcolor{gray!10}
IPR(Qwen3-Emb-0.6B) & 0.440 & 0.679 & 55.38 & 44.62 & 0.742 & 0.813 & \textbf{84.93} & 15.06\\
\rowcolor{gray!10}
IPR(Qwen3-Emb-4.B) & 0.465 & \underline{0.695} & 56.10 & 43.89 & \textbf{0.754} & \underline{0.843} & 84.25 & 15.74 \\
\bottomrule
\end{tabular}
}

\end{table*}

%% file: sec-appendix.tex
\section{Evaluation Metrics}
\label{asec:evaluation_metrics}

\subsection{Quality Prediction Metrics}
\label{asubsec:evaluation_accuracy}

Since routing decisions depend on accurate quality predictions, we validate the estimator's ranking ability:

\paragraph{Top-K Accuracy.}
Measures whether the predicted top-$k$ models match the ground-truth top-$k$ models in exact order. For $N$ candidates, we report accuracies for $k \in \{1, ..., N-1\}$.

\paragraph{Top-K F1.}
Relaxes the ordering constraint by measuring set overlap between predicted and ground-truth top-$k$ models, providing a more forgiving assessment of ranking quality.

\subsection{Routing Performance Metrics}
\label{asubsec:evaluation_routing}

\paragraph{Bounded-ARQGC.}
To evaluate routing quality across varying cost-quality trade-offs, we introduce \emph{Bounded Average Response Quality Gain under Cost} (Bounded-ARQGC). This metric generalizes the area under the quality-cost curve, normalized to $[0, 1]$.

Formally, let $Q(\alpha)$ denote the average response quality achieved when the router operates at cost budget $\alpha \cdot C_{\text{max}}$, where $C_{\text{max}}$ is the cost of always using the most expensive model. Bounded-ARQGC is defined as:
\begin{equation}
\text{Bounded-ARQGC} = \int_0^1 \frac{Q(\alpha) - Q_{\text{min}}}{Q_{\text{max}} - Q_{\text{min}}} d\alpha,
\end{equation}
where $Q_{\text{min}}$ and $Q_{\text{max}}$ are the qualities achieved by always using the cheapest and best models respectively. 
Notably, Bounded-ARQGC has following key properties:
\begin{itemize}
    \item Random routing yields $\approx 0.5$ (diagonal line).
    \item Perfect routing approaches $1.0$ (upper-left corner).
    \item Higher values indicate better cost-quality trade-offs.
\end{itemize}
Different from metrics that evaluate at fixed operating points or quality threshold values, Bounded-ARQGC captures routing performance across all possible tolerance settings, making it ideal for comparing routers without committing to specific deployment configurations.

\paragraph{Cost Save Ratio (CSR).}
For practical deployment decisions, we report cost savings at specific quality targets:
\begin{equation}
\text{CSR}(\tau) = \frac{v_{\text{best}} - v_{\text{router}}(\tau)}{v_{\text{best}}},
\end{equation}
where $v_{\text{router}}(q)$ is the cost to achieve quality level $q$ relative to the best model's quality. For instance, CSR(100\%) indicates cost savings while maintaining the best model's full quality—our primary operating point in production.

\section{Reward Modeling for Quality Supervision}
\label{asec:reward_modeling}

Training an accurate quality estimator requires large-scale supervision signals that capture human preferences over model responses. While human annotations provide the gold standard, their cost prohibits scaling to the millions of prompts needed for robust routing. We address this challenge by leveraging reward models (RMs) as automated quality evaluators.

Our approach treats response quality estimation as a regression problem: given a prompt $x_i$ and candidate model $c$'s response $y_{i,c}$, the reward model produces a quality score $r_{i,c} = \text{RM}(x_i, y_{i,c}) \in [0, 1]$. The quality estimator then learns to predict these scores directly from prompts without generating responses: $\hat{r}_{i,c} = R_\theta(x_i, c)$. 

This formulation provides three key advantages:

\textbf{Fine-grained supervision}: Unlike binary preferences or categorical labels, continuous RM scores capture subtle quality differences between models. For instance, while models may produce acceptable responses for simple queries, RMs can distinguish the incrementally better coherence or completeness that justifies routing to more capable models.

\textbf{Alignment with human judgment}: We validate that RM-based rankings align with human preferences through systematic evaluation. Model orderings derived from RM scores (e.g., Claude-3.5-Sonnet $>$ Claude-3-Opus $>$ Claude-3-Haiku) match human annotator rankings with 85\% agreement, significantly outperforming LLM-as-a-Judge approaches.

\textbf{Distribution properties}: RM scores exhibit favorable statistical properties for learning, with well-separated score distributions across models (typical separation of 0.1-0.2 between adjacent models) compared to the compressed ranges produced by LLM judges. This separation provides clearer learning signals and more stable gradient updates during training.

In practice, we employ the Skywork-Reward model~\cite{liu2024skyworkreward} to generate training labels, chosen for its strong correlation with human preferences and computational efficiency. This approach enables us to create training datasets of over 1.5M examples while maintaining quality comparable to human-annotated data.

\section{Quality Estimator Implementation Details}
\label{asec:quality_estimator_details}

\subsection{Architectural Specifications}

\textbf{Prompt Encoder Details:}
The prompt encoder uses a pretrained transformer model with fixed architecture, fine-tuned on paired prompt-score examples. For family-specific quality estimation, each model family maintains independent prompt encoders initialized from the same base encoder. Typical embedding dimension $d = 768$ for efficiency.

\textbf{LLM Identity Encoder Details:}
Learnable embeddings for each candidate model with dimension $d' = 128$. These embeddings are learned jointly with the predictor and capture model-specific behavioral patterns. For modular extension, we maintain separate LLM Identity Encoders per model family.

\textbf{Fusion Module Architecture:}
The concatenated embeddings pass through a 2-layer feed-forward network with ReLU activation:
\begin{align}
\mathbf{z}_{i,c} &= \mathrm{Concat}(\mathbf{p}_i, \mathbf{e}_c) \\
\mathbf{h} &= \mathrm{ReLU}(\mathbf{W}_1 \mathbf{z}_{i,c} + \mathbf{b}_1) \\
\hat{r}_{i,c} &= \sigma(\mathbf{W}_2 \mathbf{h} + \mathbf{b}_2)
\end{align}
where $\sigma$ is the sigmoid function to ensure output in $[0, 1]$.

\subsection{Unified vs. Family-Specific Design}

While a unified QE architecture with shared encoders and single prediction head is more compact, our experiments show superior performance using family-specific variants:
- 5-8\% higher ranking accuracy within families
- Better generalization to new models within the same family
- Simplified debugging and model-specific optimization
- Reduced interference between models with distinct output behaviors

\section{Modular Adaptation Implementation}
\label{asec:modular_adaptation}

To ensure extensibility, our design incorporates lightweight adapter modules for seamless integration of new LLMs. As illustrated in Figure~\ref{fig:ipr_model_architecture}, we freeze core encoders after initial training and attach learnable adapters for new models.

\textbf{Adapter Architecture:}
- \textbf{PE Adapter X}: 2-layer feed-forward network with residual connection, inserted after frozen prompt encoder
- \textbf{LIE Adapter X}: Single linear transformation after frozen identity encoder
- \textbf{New QP Head}: Model-specific prediction head trained from scratch

\textbf{Training Procedure:}
1. Freeze all existing model components
2. Initialize adapters with identity mapping
3. Train only adapters and new QP head on data mixture:
   - 70\% new model data
   - 30\% existing model data (for consistency)
4. Use consistency loss to maintain performance:
   \begin{equation}
   \mathcal{L} = \mathcal{L}_{\text{new}} + \lambda \sum_{i,c \in \mathcal{C}_{\text{old}}} ||\hat{r}_{i,c} - \hat{r}_{i,c}^{\text{frozen}}||^2
   \end{equation}

This framework reduces new model integration from 2-3 days of full training to 3-4 hours of adapter training, while maintaining 98\%+ performance on existing models.

\section{Human Annotation Results}
\label{asec:human_annotation}

We conducted a comprehensive evaluation of IPR-selected responses through human annotations following the specified protocol. Our evaluation framework employed a multi-batch annotation strategy to ensure robust and reliable assessments across different model families.

The human evaluation dataset was derived from a subset of the IPR test dataset.
We deliberately excluded coding-related tasks from the evaluation due to limitations in annotation expertise for technical code assessment.
The resulting dataset comprised \textbf{895 prompts}, each evaluated across \textbf{9 different models}, including 4 models from the Claude family and 5 models from the Llama family, resulting in 8055 responses.

We employed a rigorous evaluation protocol where each response underwent three blind annotation passes. The final scores were determined through majority voting across these passes, followed by calculation of the average overall satisfaction score for each model.

\paragraph{Overall Satisfaction Scores} The human annotation results revealed clear performance hierarchies within both model families. Table~\ref{tab:satisfaction_scores} presents the average overall satisfaction scores after majority voting.

\begin{table}[h]
\centering
\caption{Average Overall Satisfaction Scores by Model.}
\label{tab:satisfaction_scores}
\begin{tabular}{lc}
\toprule
\textbf{Model} & \textbf{Average Score} \\
\midrule
\multicolumn{2}{l}{\textit{Claude Family}} \\
Claude 3 Haiku & 0.8209 \\
Claude 3.5 Sonnet V1 & 0.8220 \\
Claude 3.5 Haiku & 0.8654 \\
Claude 3.5 Sonnet V2 & \textbf{0.8708} \\
\midrule
\multicolumn{2}{l}{\textit{Llama Family}} \\
Llama 3.1 8B & 0.7901 \\
Llama 3.1 70B & 0.8136 \\
Llama 3.2 11B & 0.8554 \\
Llama 3.2 90B & 0.8692 \\
Llama 3.3 70B & \textbf{0.8767} \\
\bottomrule
\end{tabular}
\end{table}

To provide more granular insights into model performance differences, we conducted pairwise comparisons for priority model pairs. Table~\ref{tab:pairwise} presents the win-tie-lose rates for three key comparisons that are critical for routing decisions.

\begin{table}[h]
\centering
\caption{Pairwise LLM Comparison Results}
\label{tab:pairwise}
\resizebox{0.5\textwidth}{!}{%
\begin{tabular}{lccc}
\toprule
\textbf{Pair} & \textbf{Win (\%)} & \textbf{Tie (\%)} & \textbf{Lose (\%)} \\
\midrule
Haiku-3 vs. Sonnet 3.5 & 11.28 & 52.85 & 31.73 \\
Haiku-3.5 vs. Sonnet 3.5 & 14.19 & 61.68 & 16.54 \\
Llama-3.2 11B vs. 3.3-70B & 12.74 & 53.18 & 20.11 \\
\bottomrule
\end{tabular}
}
\end{table}

The human annotation results demonstrate strong alignment with our expected performance hierarchies for IPR decisions. Specifically, we observe the following orderings for all priority model pairs:

\begin{enumerate}
    \item \textbf{Claude Family}: Haiku $<$ Sonnet 3.5 V2 and Haiku 3.5 $<$ Sonnet 3.5 V2
    \item \textbf{Llama Family}: Llama 3.2 11B $<$ Llama 3.3 70B
\end{enumerate}

These rankings are consistent with the reward model score comparisons, providing convergent validity for our evaluation framework. The high percentage of ties in pairwise comparisons (ranging from 52.85\% to 61.68\%) suggests that model capabilities overlap significantly for many tasks, highlighting the importance of careful model selection based on specific use case requirements.

\section{Cost Calculation Formula and Detailed Model Costs}
\label{asec:cost_computation}
We compute the routing cost as the sum of both input and output token cost per 1M tokens based on the Amazon Bedrock price list as of March 19, 2025. In our following formula, we use normalized cost to make it invariant to different datasets with different prompt or response lengths. 

Specifically, given:
\begin{itemize}
    \item A prompt $x_i$
    \item The selected LLM is $m_i$
    \item The input cost per token is $P_{m_i}$ for LLM $m_i$
    \item The output cost per token is $Q_{m_i}$ for LLM $m_i$
    \item The input prompt length is $L_{x_i}$
    \item The output response length for prompt $x_i$ and model $m_i$ is $O(x_i, m_i)$
\end{itemize}

The normalized cost for $N$ prompts and their correspondingly selected LLMs is computed as:

\begin{equation}
C = \frac{\sum_{i}^{N} L_{x_i}\times P_{m_i}}{\sum_{i}^{N} L_{x_i}}  + \frac{\sum_{i}^{N} O(x_i, m_i)\times Q_{m_i}}{\sum_{i}^{N} O(x_i, m_i)} 
\end{equation}

\subsection{Language Model Unit Prices}
Model prices for each LLM candidate is listed in Table \cref{tab:prices}. \textbf{Note:} Prices are subject to change.
\begin{table}[h]
\caption{Model pricing per 1,000 Tokens}
\label{tab:prices}
\centering
\resizebox{\columnwidth}{!}{
\begin{tabular}{l|l|c|c}
\hline
\textbf{LLM Family} & \textbf{Model} & \textbf{Input Tokens} & \textbf{Output Tokens} \\
\toprule
\multirow{4}{*}{\textbf{Anthropic}} & Claude 3.5 Sonnet V2 & \$0.003 & \$0.015 \\
& Claude 3.5 Sonnet V1 & \$0.003 & \$0.015 \\
& Claude 3.5 Haiku & \$0.0008 & \$0.004 \\
& Claude 3 Haiku & \$0.00025 & \$0.00125 \\
\hline
\multirow{5}{*}{\textbf{Llama}} & Llama 3.3 Instruct (70B) & \$0.00072 & \$0.00072 \\
& Llama 3.2 Instruct (90B) & \$0.00072 & \$0.00072 \\
& Llama 3.2 Instruct (11B) & \$0.00016 & \$0.00016 \\
& Llama 3.1 Instruct (70B) & \$0.00099 & \$0.00099 \\
& Llama 3.1 Instruct (8B) & \$0.00022 & \$0.00022 \\
\hline
\multirow{2}{*}{\textbf{Nova}} & Nova Pro & \$0.0008 & \$0.0032 \\
& Nova Lite & \$0.00006 & \$0.00024 \\
\bottomrule
\end{tabular}
}
\end{table}

\section{Dataset Collection}
\label{asec:dataset_collection}

The composition of the Combined training set is summarized in~\cref{tab:training-composition}: the largest portion comes from a multi-turn chat corpus (approximately 61\%), followed by instruction-tuning and knowledge-intensive datasets. This mixture provides broad coverage across natural language task types, allowing the quality estimator to generalize effectively across diverse prompt styles.
We determined the specific proportions in~\cref{tab:training-composition} by computing a weighted ratio for each constituent dataset, where the weight corresponds to the ratio of the original dataset size to the cumulative size of all original datasets. We then uniformly sampled from each dataset according to its assigned ratio, for instance, a dataset with proportion 60\% contributed 60\% of its datapoints to the combined training set.

The training set comprises approximately 1.5 million examples for Claude, with similar sizes for Llama and Nova after filtering out examples with response generation failures due to throttling or timeout.
Development and test sets contain between 5,000 and 6,000 examples per model family and follow a similar prompt distribution.

To evaluate generalization, we include two held-out test sets: \textbf{MS Marco}~\cite{nguyen2016msmarco} and \textbf{Nvidia Chat}~\cite{liu2024nvidiachatqa} which focus on retrieval-augmented question answering, each with around 2,000 prompts (uniformly sampled). All test responses are also scored by the \texttt{Skywork/Skywork-Reward-Gemma-2-27B} reward model to support evaluation.

\begin{table}[h]
\centering
\caption{Training dataset composition by source dataset.}
\label{tab:training-composition}
\resizebox{\linewidth}{!}{%
\begin{tabular}{lrr}
\toprule
\textbf{Dataset Name} & \textbf{Count} & \textbf{Proportion} \\
\midrule
LMSYS-Chat-1M~\cite{zheng2024lmsyschatm}        & 925,278  & 61.26\% \\
ShareGPT-Vicuna~\cite{wang2024openchat}      & 201,922  & 13.37\% \\
MixInstruct~\cite{jiang-etal-2023-llmblender}          & 98,473   & 6.52\% \\
Nectar~\cite{zhu2023starling7b}               & 98,177   & 6.50\% \\
AnswerSumm~\cite{fabbri-etal-2022-answersumm} & 42,454   & 2.81\% \\
HellaSwag~\cite{zellers-etal-2019-hellaswag}            & 41,801   & 2.77\% \\
StrategyQA~\cite{geva2021strategyqa}           & 39,385   & 2.61\% \\
CommonsenseQA~\cite{talmor-etal-2019-commonsenseqa}        & 39,081   & 2.59\% \\
BANKING77~\cite{Casanueva2020bank77}            & 14,073   & 0.93\% \\
GSM8K~\cite{cobbe2021gsm8k} & 9,771    & 0.65\% \\
\bottomrule
\end{tabular}%
}

\end{table}

\section{Ablation Studies}
\label{asec:ablation_studies}

We conduct comprehensive ablations to validate our design choices across three critical dimensions: training objectives, architectural decisions, and routing strategies.
\begin{table}[h]
\centering
\caption{Comparison of training loss functions (averaged over three model families).}
\label{tab:loss_comparison}
\resizebox{\columnwidth}{!}{%
\begin{tabular}{l|c|c|c|c}
\hline
\textbf{Loss} & \textbf{B-ARQGC} & \textbf{Quality} & \textbf{CSR} & \textbf{Route Acc} \\
\hline
MSE        & 0.7361 & 0.5451 & 0.3130 & 0.6353 \\
Hinge Loss & 0.6897 & 0.5438 & 0.2660 & 0.6035 \\
ListNet    & 0.7292 & 0.5448 & 0.2656 & 0.5673 \\
\hline
\end{tabular}%
}

\end{table}

\paragraph{Training Loss Functions.}
Table~\ref{tab:loss_comparison} compares different loss functions for training the quality estimator, averaged across all model families. While we experimented with ranking-based losses that directly optimize for relative ordering, MSE loss achieves the best overall performance with 0.736 Bounded-ARQGC, outperforming hinge loss by 6.7\% and ListNet by 0.9\%. 
This result can be explained by two factors. First, continuous regression targets provide richer gradient signals than pairwise or listwise comparisons, enabling more stable optimization. Second, MSE loss naturally captures the magnitude of quality differences, which proves crucial for threshold-based routing decisions. Interestingly, while hinge loss achieves comparable routing accuracy (60.3\% vs 63.5\%), it significantly underperforms in cost savings (26.6\% vs 31.3\%), suggesting that accurate quality magnitude estimation is more important than perfect ranking for cost-optimal routing.

\begin{figure*}[ptb]
    \centering
    \includegraphics[width=0.9\linewidth]{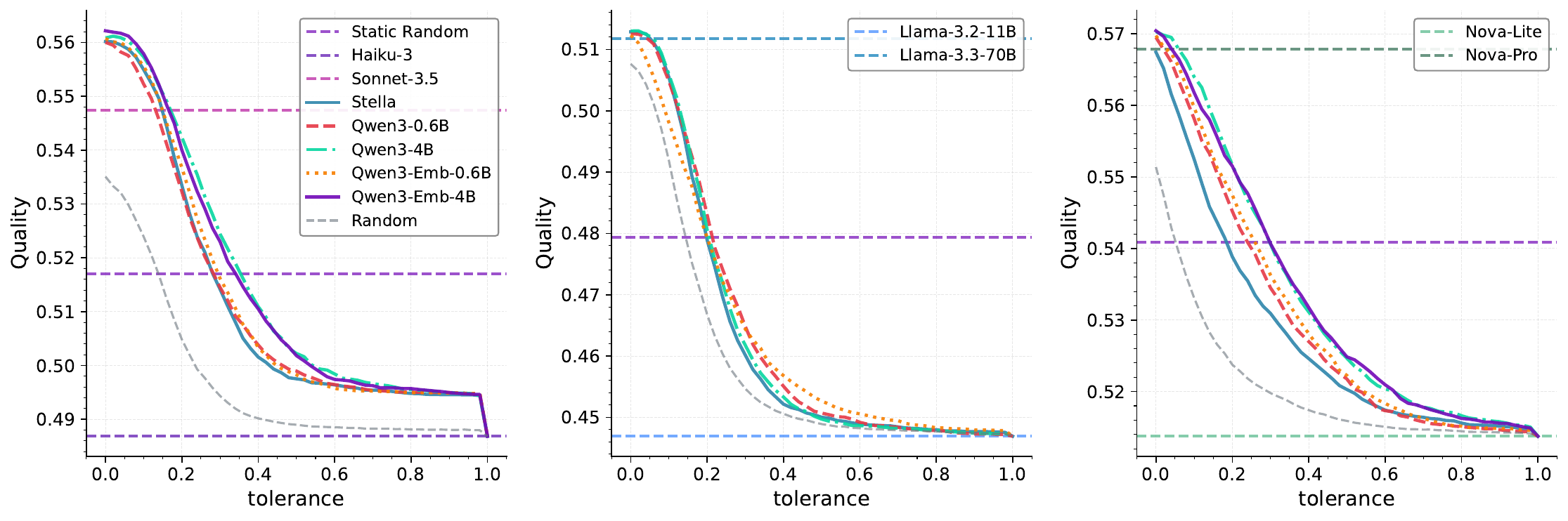}
    \caption{Quality v.s. tolerance with different QE backbones.}
    \vspace{-10pt}
    \label{fig:quality_tol_quality_backbone}
\end{figure*}

\begin{figure*}[ptb]
    \centering
    \includegraphics[width=0.95\linewidth]{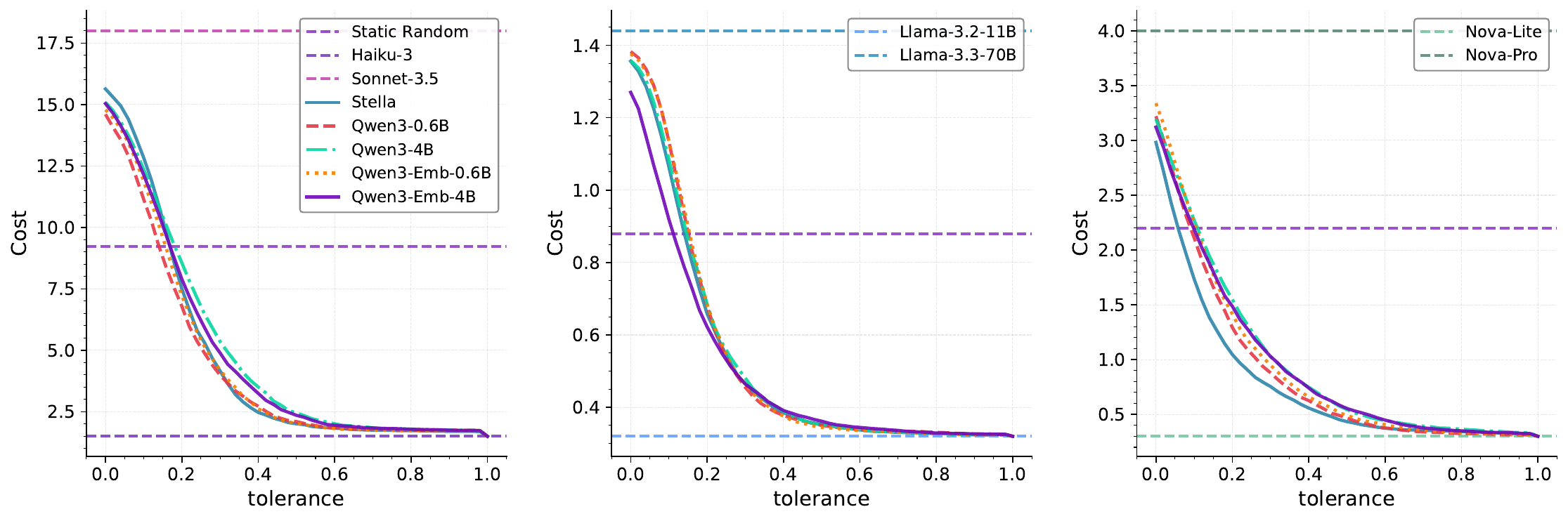}
    \caption{Cost v.s. tolerance with different QE backbones.}
    \vspace{-10pt}
    \label{fig:quality_tol_cost_backbone}
\end{figure*}

\paragraph{Family-Specific vs. Unified Routing.}
Table~\ref{tab:model_comparison} examines the trade-offs between training separate routers for each model family versus a single unified router. Family-specific routers consistently outperform unified approaches on in-domain data, achieving higher Bounded-ARQGC scores (0.799 vs 0.792 for Claude, 0.663 vs 0.659 for Llama, and 0.731 vs 0.729 for Nova). This specialization advantage stems from the reduced learning complexity\,---\,each router only needs to focus on quality patterns within a homogeneous model group. Conversely, unified routers excel at generalization: on out-of-distribution datasets, they achieve 5.7\%, 1.6\%, and 7.6\% higher Bounded-ARQGC for Claude, Llama, and Nova respectively. This reveals a bias-variance trade-off where family-specific routers precisely capture in-domain patterns but may overfit to family-specific characteristics. Given our production emphasis on in-domain performance, we deploy family-specific routing while recognizing unified routing's merits for heterogeneous workloads.
\begin{table*}[h]
\centering
\caption{In- and out-of-distribution performance comparison of family-specific and unified routers. Cost-saving ratio (CSR) and routing accuracy (ACC) are reported at 100\% best candidate performance. Bold values indicate superior performance within each distribution type.}
\label{tab:model_comparison}

\resizebox{0.9\textwidth}{!}{%
\begin{tabular}{l|l|c|c|c|c||c|c|c|c}
\toprule
& & \multicolumn{4}{c||}{\textbf{In-Distribution}} & \multicolumn{4}{c}{\textbf{Out-of-Distribution}} \\
\cmidrule(lr){3-6} \cmidrule(lr){7-10}
\textbf{Model} & \textbf{Type} & \textbf{MAE} $\downarrow$ & \textbf{B-ARQGC} $\uparrow$ & \textbf{CSR} $\uparrow$ & \textbf{ACC} $\uparrow$ & \textbf{MAE} $\downarrow$ & \textbf{B-ARQGC} $\uparrow$ & \textbf{CSR} $\uparrow$ & \textbf{ACC} $\uparrow$ \\
\midrule
\multirow{2}{*}{Claude} & \cellcolor{gray!15}specific & \cellcolor{gray!15}0.09478 & \cellcolor{gray!15}\textbf{0.799} & \cellcolor{gray!15}\textbf{0.439} & \cellcolor{gray!15}\textbf{0.678} & \cellcolor{gray!15}0.1532 & \cellcolor{gray!15}0.523 & \cellcolor{gray!15}0.369 & \cellcolor{gray!15}0.57 \\
\cline{2-10}
& unified & 0.1005 & 0.792 & 0.421 & 0.668 & \textbf{0.142} & \textbf{0.553} & \textbf{0.398} & \textbf{0.61} \\
\midrule
\multirow{2}{*}{Llama} & \cellcolor{gray!15}specific & \cellcolor{gray!15}\textbf{0.08626} & \cellcolor{gray!15}\textbf{0.663} & \cellcolor{gray!15}\textbf{0.0773} & \cellcolor{gray!15}\textbf{0.677} & \cellcolor{gray!15}0.1221 & \cellcolor{gray!15}0.512 & \cellcolor{gray!15}0.0712 & \cellcolor{gray!15}0.59 \\
\cline{2-10}
& unified & 0.08710 & 0.659 & 0.0720 & 0.672 & \textbf{0.1190} & \textbf{0.520} & \textbf{0.0725} & \textbf{0.60} \\
\midrule
\multirow{2}{*}{Nova} & \cellcolor{gray!15}specific & \cellcolor{gray!15}\textbf{0.09597} & \cellcolor{gray!15}\textbf{0.731} & \cellcolor{gray!15}\textbf{0.255} & \cellcolor{gray!15}\textbf{0.652} & \cellcolor{gray!15}0.1447 & \cellcolor{gray!15}0.525 & \cellcolor{gray!15}0.152 & \cellcolor{gray!15}0.60 \\
\cline{2-10}
& unified & 0.1021 & 0.729 & 0.242 & 0.648 & \textbf{0.1324} & \textbf{0.565} & \textbf{0.180} & \textbf{0.64} \\
\bottomrule
\end{tabular}%
}

\end{table*}

\begin{table}[h]
\centering
\caption{Routing strategy comparison.}
\label{tab:routing-strategy}
\begin{tabular}{lcc}
\hline
\textbf{Strategy} & \textbf{min} & \textbf{max} \\
\hline
dynamic max      & 0       & dynamic \\
dynamic minmax   & dynamic & dynamic \\
static dynamic   & static  & dynamic \\
static           & static  & static \\
\hline
\end{tabular}

\end{table}

\begin{figure*}[h!]
  \centering
  \includegraphics[width=0.33\textwidth]{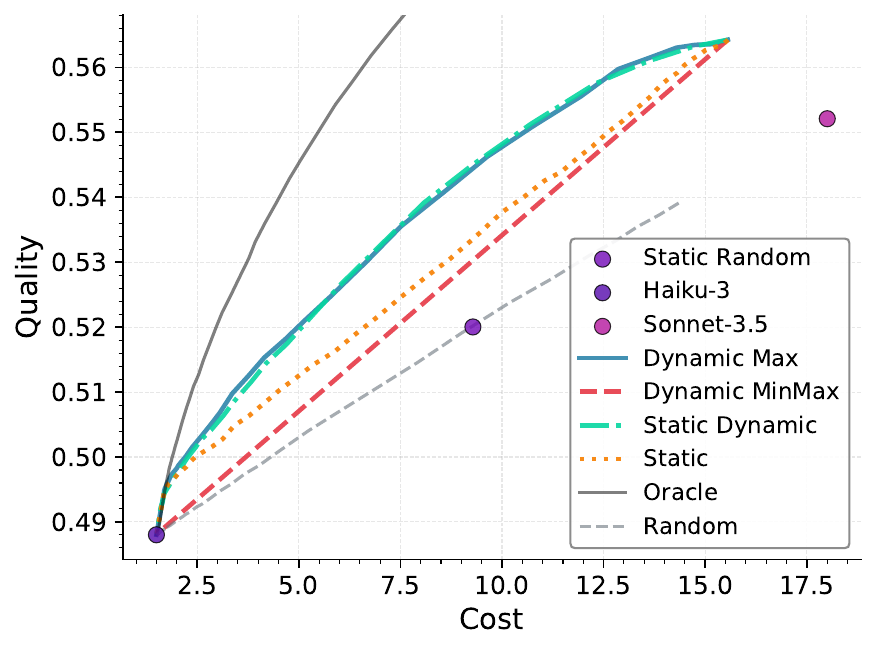}\hfill
  \includegraphics[width=0.33\textwidth]{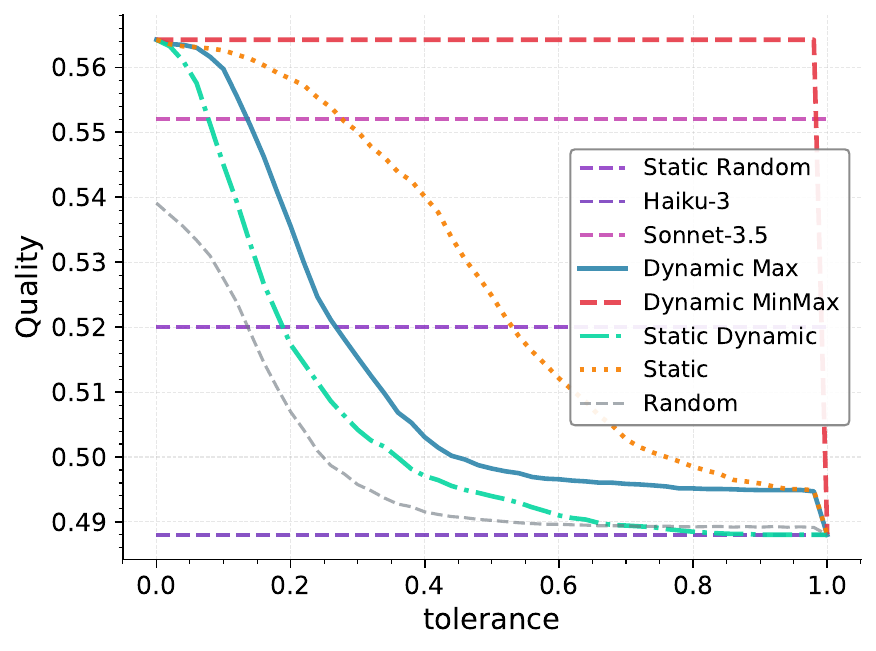}\hfill
  \includegraphics[width=0.33\textwidth]{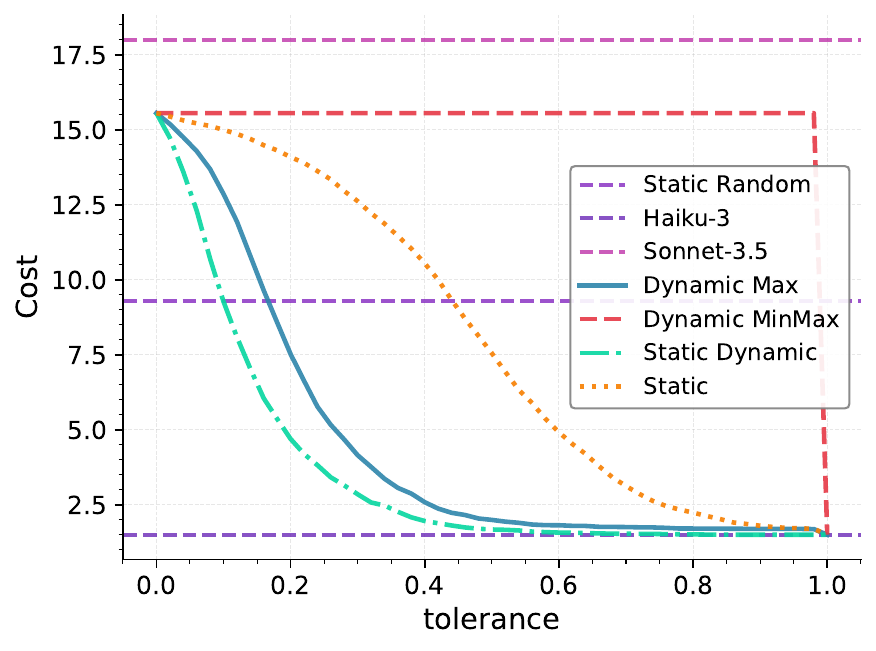}
  \caption{Quality-performance trade-off (left), quality-tolerance (middle), and cost-tolerance(right) relationship with different routing strategies}
  \label{fig:routing_strategy}
\end{figure*}

\paragraph{Routing Strategy and Threshold Calibration.}
We ablate two key aspects of our routing algorithm: the threshold computation method (dynamic vs. static) and the quality reference point.
As described in Section~\ref{subsec:routing}, dynamic thresholds adapt to each prompt's quality distribution while static thresholds use global statistics.
As shown in Figure \ref{fig:routing_strategy}, our experiments reveal that Dynamic Max and Dynamic MinMax achieves the optimal AUC compares to others.
Among this two, Dynamic Max has more smooth quality and cost curve vs tolerance compared to Dynamic MinMax, giving user more freedom to control the routing behavior. 
This hybrid strategy effectively handles prompts with varying difficulty\,---\,easy prompts with high quality scores across all models benefit from dynamic adaptation, while the fixed minimum prevents threshold collapse for uniformly challenging prompts. 
This also indicates the per-prompt normalization is crucial: without adapting thresholds to individual quality distributions, routers exhibit excessive conservatism, routing more prompts to expensive models unnecessarily. 

\section{Extended Discussions on Related Works}
\label{asec:relatedworks}

\paragraph{Routing benchmarks.}
Existing LLM routing benchmarks are mostly curated from popular NLP datasets covering different facets of LLM usage. MixInstruct~\cite{jiang-etal-2023-llmblender} consists of 110k examples focusing on the chat capability of LLMs. The mixture is primarily from four sources: Alpaca-GPT4~\cite{stanfordalpaca}, Dolly-15K~\cite{DatabricksBlog2023DollyV2}, GPT4All-LAION~\cite{gpt4laion} and ShareGPT~\cite{wang2024openchat}.
RouterBench~\cite{hu2024routerbench} constructs a benchmark with over 405k inference outcomes from 11 representative LLMs across 8 diverse datasets to support the development of routing strategies. Routing strategies covered in RouterBench are simple methods like KNN and MLP routers.
RouterEval~\cite{huang2025routereval} is a concurrent work that curates a large scale evaluation benchmark, spanning 12 popular LLM evaluations across various areas such as commonsense reasoning, semantic understanding, etc, and including over 200M performance records. 
This technical report describes our solution to curate IPR dataset, an industrial-scale LLM routing benchmark that focuses on natural language understanding and text generation capabilities of LLMs, and includes models currently served on our platform.

\paragraph{Routing evaluations.} 
Existing works~\cite[\interalia]{ong2024routellm,hu2024routerbench,huang2025routereval,lu-etal-2024-routingtotheexpert} mostly categorize evaluation metrics into two groups: (1) effectiveness metrics and (2) efficiency metrics.\footnote{Some works like~\cite{huang2025routereval} also refer to as (1) routing performance metrics and (2) cost reduction metrics.}

Effectiveness metrics directly evaluate measure whether a query is routed to the most performant routing candidates. RouterEval~\cite{huang2025routereval}, CP-Router~\cite{su2025cprouter} and Self-REF~\cite{chuang2025learningtoroutellmswithconfidencetokens} evaluate routing effectiveness by Accuracy, i.e., the correctness of final predictions, which can be considered a top-1 metric. HybridLLM~\cite{ding2024hybridllmcostefficientandqualityawarequeryrouting} focus on text generation tasks and adopt BARTScore~\cite{yuan2021bartscore} as the quality/effectiveness metric. RouteLLM~\cite{ong2024routellm} defines an \emph{average response quality} score that covers different NLP tasks, e.g., correctness on golden-labeled dataset or a numerical rating. In this technical report, we focus on top-1 accuracy, F1 scores as well as AUC as the main performance metrics.

Different from straightforward effectiveness metrics, there lacks a established efficiency metric that applies to different models and platforms, due to different notions of cost definitions. For example, HybridLLM~\cite{ding2024hybridllmcostefficientandqualityawarequeryrouting} directly use the monetary cost as a proxy for the cost metric, e.g., \$ per 1M tokens. Some works like~\cite{su2025cprouter} uses number of tokens to represent the cost.
In contrast to this absolute cost metric, works such as RouteLLM~\cite{ong2024routellm} adopt relative cost efficiency metric. For example, RouteLLM~\cite{ong2024routellm} define the cost efficiency metric as the percentage of calls to strong models. 
For evaluation of IPR, we adopt the proposed Bounded Average Response Quality Gain under Cost and Cost Save Ratio as the main efficiency metrics.
We should note that, due to quick advancement of LLM inference optimization, exemplified by frameworks like vLLM~\cite{kwon2023pagedattention} and SGLang~\cite{zheng2024sglang}, the cost metric needs to be actively refreshed to reflect the actual inference cost.

The central goal of the LLM routing problem is to optimize the trade-off between effectiveness and efficiency. Various metrics have been used for evaluation and subsequently adopted as training objectives for the Router.
We skip the detailed discussions and kindly refer readers to those original works for design rationales and exact formulations.

%% file: acl.bbl
\begin{thebibliography}{43}
\providecommand{\natexlab}[1]{#1}

\bibitem[{Burges(2010)}]{burges2010ranknet}
Christopher~JC Burges. 2010.
\newblock From ranknet to lambdarank to lambdamart: An overview.
\newblock \emph{Learning}, 11(23-581):81.

\bibitem[{Casanueva et~al.(2020)Casanueva, Temcinas, Gerz, Henderson, and Vulic}]{Casanueva2020bank77}
I{\~{n}}igo Casanueva, Tadas Temcinas, Daniela Gerz, Matthew Henderson, and Ivan Vulic. 2020.
\newblock \href {https://arxiv.org/abs/2003.04807} {Efficient intent detection with dual sentence encoders}.
\newblock In \emph{Proceedings of the 2nd Workshop on NLP for ConvAI - ACL 2020}.
\newblock Data available at https://github.com/PolyAI-LDN/task-specific-datasets.

\bibitem[{Chen et~al.(2023)Chen, Zaharia, and Zou}]{chen2023frugalgpt}
Lingjiao Chen, Matei Zaharia, and James Zou. 2023.
\newblock Frugalgpt: How to use large language models while reducing cost and improving performance.
\newblock \emph{arXiv preprint arXiv:2305.05176}.

\bibitem[{Chen et~al.(2024)Chen, Jiang, Lin, Kwok, and Zhang}]{chen2024routerdc}
Shuhao Chen, Weisen Jiang, Baijiong Lin, James Kwok, and Yu~Zhang. 2024.
\newblock Routerdc: Query-based router by dual contrastive learning for assembling large language models.
\newblock \emph{Advances in Neural Information Processing Systems}, 37:66305--66328.

\bibitem[{Chuang et~al.(2025)Chuang, Sarma, Gopalan, Boccio, Bolouki, Hu, and Zhou}]{chuang2025learningtoroutellmswithconfidencetokens}
Yu-Neng Chuang, Prathusha~Kameswara Sarma, Parikshit Gopalan, John Boccio, Sara Bolouki, Xia Hu, and Helen Zhou. 2025.
\newblock \href {https://openreview.net/forum?id=U08mUogGDM} {Learning to route {LLM}s with confidence tokens}.
\newblock In \emph{Forty-second International Conference on Machine Learning}.

\bibitem[{Cobbe et~al.(2021)Cobbe, Kosaraju, Bavarian, Chen, Jun, Kaiser, Plappert, Tworek, Hilton, Nakano et~al.}]{cobbe2021gsm8k}
Karl Cobbe, Vineet Kosaraju, Mohammad Bavarian, Mark Chen, Heewoo Jun, Lukasz Kaiser, Matthias Plappert, Jerry Tworek, Jacob Hilton, Reiichiro Nakano, and 1 others. 2021.
\newblock Training verifiers to solve math word problems.
\newblock \emph{arXiv preprint arXiv:2110.14168}.

\bibitem[{Conover et~al.(2023)Conover, Hayes, Mathur, Xie, Wan, Shah, Ghodsi, Wendell, Zaharia, and Xin}]{DatabricksBlog2023DollyV2}
Mike Conover, Matt Hayes, Ankit Mathur, Jianwei Xie, Jun Wan, Sam Shah, Ali Ghodsi, Patrick Wendell, Matei Zaharia, and Reynold Xin. 2023.
\newblock Free dolly: Introducing the world's first truly open instruction-tuned llm.

\bibitem[{Ding et~al.(2024)Ding, Mallick, Wang, Sim, Mukherjee, R{\"u}hle, Lakshmanan, and Awadallah}]{ding2024hybridllmcostefficientandqualityawarequeryrouting}
Dujian Ding, Ankur Mallick, Chi Wang, Robert Sim, Subhabrata Mukherjee, Victor R{\"u}hle, Laks V.~S. Lakshmanan, and Ahmed~Hassan Awadallah. 2024.
\newblock \href {https://openreview.net/forum?id=02f3mUtqnM} {Hybrid {LLM}: Cost-efficient and quality-aware query routing}.
\newblock In \emph{The Twelfth International Conference on Learning Representations}.

\bibitem[{Ding et~al.(2025)Ding, Mallick, Zhang, Wang, Madrigal, Garcia, Xia, Lakshmanan, Wu, and R{\"u}hle}]{ding2025bestroute}
Dujian Ding, Ankur Mallick, Shaokun Zhang, Chi Wang, Daniel Madrigal, Mirian Del Carmen~Hipolito Garcia, Menglin Xia, Laks V.~S. Lakshmanan, Qingyun Wu, and Victor R{\"u}hle. 2025.
\newblock \href {https://openreview.net/forum?id=tFBIbCVXkG} {{BEST}-route: Adaptive {LLM} routing with test-time optimal compute}.
\newblock In \emph{Forty-second International Conference on Machine Learning}.

\bibitem[{Fabbri et~al.(2022)Fabbri, Wu, Iyer, Li, and Diab}]{fabbri-etal-2022-answersumm}
Alexander Fabbri, Xiaojian Wu, Srini Iyer, Haoran Li, and Mona Diab. 2022.
\newblock \href {https://doi.org/10.18653/v1/2022.naacl-main.180} {{A}nswer{S}umm: A manually-curated dataset and pipeline for answer summarization}.
\newblock In \emph{Proceedings of the 2022 Conference of the North American Chapter of the Association for Computational Linguistics: Human Language Technologies}, pages 2508--2520, Seattle, United States. Association for Computational Linguistics.

\bibitem[{Feng et~al.(2025)Feng, Shen, and You}]{feng2025graphrouter}
Tao Feng, Yanzhen Shen, and Jiaxuan You. 2025.
\newblock \href {https://openreview.net/forum?id=eU39PDsZtT} {Graphrouter: A graph-based router for {LLM} selections}.
\newblock In \emph{The Thirteenth International Conference on Learning Representations}.

\bibitem[{Geva et~al.(2021)Geva, Khashabi, Segal, Khot, Roth, and Berant}]{geva2021strategyqa}
Mor Geva, Daniel Khashabi, Elad Segal, Tushar Khot, Dan Roth, and Jonathan Berant. 2021.
\newblock Did aristotle use a laptop? a question answering benchmark with implicit reasoning strategies.
\newblock \emph{Transactions of the Association for Computational Linguistics}, 9:346--361.

\bibitem[{Guo et~al.(2025)Guo, Yang, Zhang, Song, Zhang, Xu, Zhu, Ma, Wang, Bi et~al.}]{guo2025deepseek}
Daya Guo, Dejian Yang, Haowei Zhang, Junxiao Song, Ruoyu Zhang, Runxin Xu, Qihao Zhu, Shirong Ma, Peiyi Wang, Xiao Bi, and 1 others. 2025.
\newblock Deepseek-r1: Incentivizing reasoning capability in llms via reinforcement learning.
\newblock \emph{arXiv preprint arXiv:2501.12948}.

\bibitem[{Hu et~al.(2024)Hu, Bieker, Li, Jiang, Keigwin, Ranganath, Keutzer, and Upadhyay}]{hu2024routerbench}
Qitian~Jason Hu, Jacob Bieker, Xiuyu Li, Nan Jiang, Benjamin Keigwin, Gaurav Ranganath, Kurt Keutzer, and Shriyash~Kaustubh Upadhyay. 2024.
\newblock \href {https://openreview.net/forum?id=IVXmV8Uxwh} {Routerbench: A benchmark for multi-{LLM} routing system}.
\newblock In \emph{Agentic Markets Workshop at ICML 2024}.

\bibitem[{Huang et~al.(2025)Huang, Ling, Lin, Chen, Zhong, Wu, and Lin}]{huang2025routereval}
Zhongzhan Huang, Guoming Ling, Yupei Lin, Yandong Chen, Shanshan Zhong, Hefeng Wu, and Liang Lin. 2025.
\newblock Routereval: A comprehensive benchmark for routing llms to explore model-level scaling up in llms.
\newblock \emph{arXiv preprint arXiv:2503.10657}.

\bibitem[{Jaech et~al.(2024)Jaech, Kalai, Lerer, Richardson, El-Kishky, Low, Helyar, Madry, Beutel, Carney et~al.}]{jaech2024openaio1systemcard}
Aaron Jaech, Adam Kalai, Adam Lerer, Adam Richardson, Ahmed El-Kishky, Aiden Low, Alec Helyar, Aleksander Madry, Alex Beutel, Alex Carney, and 1 others. 2024.
\newblock Openai o1 system card.
\newblock \emph{arXiv preprint arXiv:2412.16720}.

\bibitem[{Jiang et~al.(2023)Jiang, Ren, and Lin}]{jiang-etal-2023-llmblender}
Dongfu Jiang, Xiang Ren, and Bill~Yuchen Lin. 2023.
\newblock \href {https://doi.org/10.18653/v1/2023.acl-long.792} {{LLM}-blender: Ensembling large language models with pairwise ranking and generative fusion}.
\newblock In \emph{Proceedings of the 61st Annual Meeting of the Association for Computational Linguistics (Volume 1: Long Papers)}, pages 14165--14178, Toronto, Canada. Association for Computational Linguistics.

\bibitem[{Jin et~al.(2025)Jin, Shao, Wen, Wu, Feng, Zhang, and Tao}]{jin2025radialrouter}
Ruihan Jin, Pengpeng Shao, Zhengqi Wen, Jinyang Wu, Mingkuan Feng, Shuai Zhang, and Jianhua Tao. 2025.
\newblock Radialrouter: Structured representation for efficient and robust large language models routing.
\newblock \emph{arXiv preprint arXiv:2506.03880}.

\bibitem[{Jitkrittum et~al.(2025)Jitkrittum, Narasimhan, Rawat, Juneja, Wang, Lee, Shenoy, Panigrahy, Menon, and Kumar}]{jitkrittum2025universalmodelroutingforefficientllminference}
Wittawat Jitkrittum, Harikrishna Narasimhan, Ankit~Singh Rawat, Jeevesh Juneja, Zifeng Wang, Chen-Yu Lee, Pradeep Shenoy, Rina Panigrahy, Aditya~Krishna Menon, and Sanjiv Kumar. 2025.
\newblock Universal model routing for efficient llm inference.
\newblock \emph{arXiv preprint arXiv:2502.08773}.

\bibitem[{Kwon et~al.(2023)Kwon, Li, Zhuang, Sheng, Zheng, Yu, Gonzalez, Zhang, and Stoica}]{kwon2023pagedattention}
Woosuk Kwon, Zhuohan Li, Siyuan Zhuang, Ying Sheng, Lianmin Zheng, Cody~Hao Yu, Joseph Gonzalez, Hao Zhang, and Ion Stoica. 2023.
\newblock Efficient memory management for large language model serving with pagedattention.
\newblock In \emph{Proceedings of the 29th Symposium on Operating Systems Principles}, pages 611--626.

\bibitem[{LAION-AI(2023)}]{gpt4laion}
LAION-AI. 2023.
\newblock Laion-ai open assistant.
\newblock \url{https://github.com/LAION-AI/Open-Assistant}.

\bibitem[{Liu et~al.(2024{\natexlab{a}})Liu, Zeng, Liu, Yan, He, Wang, Yan, Liu, and Zhou}]{liu2024skyworkreward}
Chris~Yuhao Liu, Liang Zeng, Jiacai Liu, Rui Yan, Jujie He, Chaojie Wang, Shuicheng Yan, Yang Liu, and Yahui Zhou. 2024{\natexlab{a}}.
\newblock Skywork-reward: Bag of tricks for reward modeling in llms.
\newblock \emph{arXiv preprint arXiv:2410.18451}.

\bibitem[{Liu et~al.(2024{\natexlab{b}})Liu, Ping, Roy, Xu, Lee, Shoeybi, and Catanzaro}]{liu2024nvidiachatqa}
Zihan Liu, Wei Ping, Rajarshi Roy, Peng Xu, Chankyu Lee, Mohammad Shoeybi, and Bryan Catanzaro. 2024{\natexlab{b}}.
\newblock Chatqa: Surpassing gpt-4 on conversational qa and rag.
\newblock \emph{arXiv preprint arXiv:2401.10225}.

\bibitem[{Lu et~al.(2024)Lu, Yuan, Lin, Lin, Yuan, Zhou, and Zhou}]{lu-etal-2024-routingtotheexpert}
Keming Lu, Hongyi Yuan, Runji Lin, Junyang Lin, Zheng Yuan, Chang Zhou, and Jingren Zhou. 2024.
\newblock \href {https://doi.org/10.18653/v1/2024.naacl-long.109} {Routing to the expert: Efficient reward-guided ensemble of large language models}.
\newblock In \emph{Proceedings of the 2024 Conference of the North American Chapter of the Association for Computational Linguistics: Human Language Technologies (Volume 1: Long Papers)}, pages 1964--1974, Mexico City, Mexico. Association for Computational Linguistics.

\bibitem[{Mei et~al.(2025)Mei, Xu, Lin, and Zhang}]{mei2025omnirouter}
Kai Mei, Wujiang Xu, Shuhang Lin, and Yongfeng Zhang. 2025.
\newblock Omnirouter: Budget and performance controllable multi-llm routing.
\newblock \emph{arXiv preprint arXiv:2502.20576}.

\bibitem[{Nguyen et~al.(2016)Nguyen, Rosenberg, Song, Gao, Tiwary, Majumder, and Deng}]{nguyen2016msmarco}
Tri Nguyen, Mir Rosenberg, Xia Song, Jianfeng Gao, Saurabh Tiwary, Rangan Majumder, and Li~Deng. 2016.
\newblock \href {https://arxiv.org/abs/1611.09268} {{MS} {MARCO:} {A} human generated machine reading comprehension dataset}.
\newblock \emph{CoRR}, abs/1611.09268.

\bibitem[{Ong et~al.(2024)Ong, Almahairi, Wu, Chiang, Wu, Gonzalez, Kadous, and Stoica}]{ong2024routellm}
Isaac Ong, Amjad Almahairi, Vincent Wu, Wei-Lin Chiang, Tianhao Wu, Joseph~E Gonzalez, M~Waleed Kadous, and Ion Stoica. 2024.
\newblock Routellm: Learning to route llms from preference data.
\newblock In \emph{The Thirteenth International Conference on Learning Representations}.

\bibitem[{Pan et~al.(2025)Pan, Zhang, Zhao, and Han}]{pan2025routetoreason}
Zhihong Pan, Kai Zhang, Yuze Zhao, and Yupeng Han. 2025.
\newblock Route to reason: Adaptive routing for llm and reasoning strategy selection.
\newblock \emph{arXiv preprint arXiv:2505.19435}.

\bibitem[{Sakota et~al.(2024)Sakota, Peyrard, and West}]{sakota2024forc}
Marija Sakota, Maxime Peyrard, and Robert West. 2024.
\newblock \href {https://doi.org/10.1145/3616855.3635825} {Fly-swat or cannon? cost-effective language model choice via meta-modeling}.
\newblock In \emph{Proceedings of the 17th ACM International Conference on Web Search and Data Mining}, WSDM '24, page 606–615, New York, NY, USA. Association for Computing Machinery.

\bibitem[{Sikeridis et~al.(2024)Sikeridis, Ramdass, and Pareek}]{sikeridis2024pickllm}
Dimitrios Sikeridis, Dennis Ramdass, and Pranay Pareek. 2024.
\newblock Pickllm: Context-aware rl-assisted large language model routing.
\newblock \emph{arXiv preprint arXiv:2412.12170}.

\bibitem[{Sikeridis et~al.(2025)Sikeridis, Ramdass, and Pareek}]{sikeridis2025pickllm}
Dimitrios Sikeridis, Dennis Ramdass, and Pranay Pareek. 2025.
\newblock Pickllm: Context-aware rl-assisted large language model routing.
\newblock In \emph{International Workshop on AI for Transportation}, pages 227--239. Springer.

\bibitem[{Stripelis et~al.(2024)Stripelis, Hu, Zhang, Xu, Shah, Jin, Yao, Avestimehr, and He}]{stripelis2024tensoropera}
Dimitris Stripelis, Zijian Hu, Jipeng Zhang, Zhaozhuo Xu, Alay~Dilipbhai Shah, Han Jin, Yuhang Yao, Salman Avestimehr, and Chaoyang He. 2024.
\newblock Tensoropera router: A multi-model router for efficient llm inference.
\newblock \emph{arXiv preprint arXiv:2408.12320}.

\bibitem[{Su et~al.(2025)Su, Lin, Feng, Zheng, Wang, Xiao, Zhao, Liu, Cheng, and Wang}]{su2025cprouter}
Jiayuan Su, Fulin Lin, Zhaopeng Feng, Han Zheng, Teng Wang, Zhenyu Xiao, Xinlong Zhao, Zuozhu Liu, Lu~Cheng, and Hongwei Wang. 2025.
\newblock Cp-router: An uncertainty-aware router between llm and lrm.
\newblock \emph{arXiv preprint arXiv:2505.19970}.

\bibitem[{Talmor et~al.(2019)Talmor, Herzig, Lourie, and Berant}]{talmor-etal-2019-commonsenseqa}
Alon Talmor, Jonathan Herzig, Nicholas Lourie, and Jonathan Berant. 2019.
\newblock \href {https://doi.org/10.18653/v1/N19-1421} {{C}ommonsense{QA}: A question answering challenge targeting commonsense knowledge}.
\newblock In \emph{Proceedings of the 2019 Conference of the North {A}merican Chapter of the Association for Computational Linguistics: Human Language Technologies, Volume 1 (Long and Short Papers)}, pages 4149--4158, Minneapolis, Minnesota. Association for Computational Linguistics.

\bibitem[{Taori et~al.(2023)Taori, Gulrajani, Zhang, Dubois, Li, Guestrin, Liang, and Hashimoto}]{stanfordalpaca}
Rohan Taori, Ishaan Gulrajani, Tianyi Zhang, Yann Dubois, Xuechen Li, Carlos Guestrin, Percy Liang, and Tatsunori~B. Hashimoto. 2023.
\newblock Stanford alpaca: An instruction-following llama model.
\newblock \url{https://github.com/tatsu-lab/stanford_alpaca}.

\bibitem[{Wang et~al.(2024)Wang, Cheng, Zhan, Li, Song, and Liu}]{wang2024openchat}
Guan Wang, Sijie Cheng, Xianyuan Zhan, Xiangang Li, Sen Song, and Yang Liu. 2024.
\newblock \href {https://openreview.net/forum?id=AOJyfhWYHf} {Openchat: Advancing open-source language models with mixed-quality data}.
\newblock In \emph{The Twelfth International Conference on Learning Representations}.

\bibitem[{Yuan et~al.(2021)Yuan, Neubig, and Liu}]{yuan2021bartscore}
Weizhe Yuan, Graham Neubig, and Pengfei Liu. 2021.
\newblock Bartscore: Evaluating generated text as text generation.
\newblock \emph{Advances in neural information processing systems}, 34:27263--27277.

\bibitem[{Yue et~al.(2024)Yue, Zhao, Zhang, Du, and Yao}]{yue2024largelanguagemodelcascades}
Murong Yue, Jie Zhao, Min Zhang, Liang Du, and Ziyu Yao. 2024.
\newblock \href {https://openreview.net/forum?id=6okaSfANzh} {Large language model cascades with mixture of thought representations for cost-efficient reasoning}.
\newblock In \emph{The Twelfth International Conference on Learning Representations}.

\bibitem[{Zellers et~al.(2019)Zellers, Holtzman, Bisk, Farhadi, and Choi}]{zellers-etal-2019-hellaswag}
Rowan Zellers, Ari Holtzman, Yonatan Bisk, Ali Farhadi, and Yejin Choi. 2019.
\newblock \href {https://doi.org/10.18653/v1/P19-1472} {{H}ella{S}wag: Can a machine really finish your sentence?}
\newblock In \emph{Proceedings of the 57th Annual Meeting of the Association for Computational Linguistics}, pages 4791--4800, Florence, Italy. Association for Computational Linguistics.

\bibitem[{Zheng et~al.(2024{\natexlab{a}})Zheng, Chiang, Sheng, Li, Zhuang, Wu, Zhuang, Li, Lin, Xing, Gonzalez, Stoica, and Zhang}]{zheng2024lmsyschatm}
Lianmin Zheng, Wei-Lin Chiang, Ying Sheng, Tianle Li, Siyuan Zhuang, Zhanghao Wu, Yonghao Zhuang, Zhuohan Li, Zi~Lin, Eric Xing, Joseph~E. Gonzalez, Ion Stoica, and Hao Zhang. 2024{\natexlab{a}}.
\newblock \href {https://openreview.net/forum?id=BOfDKxfwt0} {{LMSYS}-chat-1m: A large-scale real-world {LLM} conversation dataset}.
\newblock In \emph{The Twelfth International Conference on Learning Representations}.

\bibitem[{Zheng et~al.(2024{\natexlab{b}})Zheng, Yin, Xie, Sun, Huang, Yu, Cao, Kozyrakis, Stoica, Gonzalez, Barrett, and Sheng}]{zheng2024sglang}
Lianmin Zheng, Liangsheng Yin, Zhiqiang Xie, Chuyue Sun, Jeff Huang, Cody~Hao Yu, Shiyi Cao, Christos Kozyrakis, Ion Stoica, Joseph~E. Gonzalez, Clark Barrett, and Ying Sheng. 2024{\natexlab{b}}.
\newblock \href {https://openreview.net/forum?id=VqkAKQibpq} {{SGL}ang: Efficient execution of structured language model programs}.
\newblock In \emph{The Thirty-eighth Annual Conference on Neural Information Processing Systems}.

\bibitem[{Zhu et~al.(2023)Zhu, Frick, Wu, Zhu, and Jiao}]{zhu2023starling7b}
Banghua Zhu, Evan Frick, Tianhao Wu, Hanlin Zhu, and Jiantao Jiao. 2023.
\newblock Starling-7b: Improving llm helpfulness \& harmlessness with rlaif.

\bibitem[{Zhuang et~al.(2024)Zhuang, Wu, Wen, Li, Jiao, and Ramchandran}]{zhuang2024embedllm}
Richard Zhuang, Tianhao Wu, Zhaojin Wen, Andrew Li, Jiantao Jiao, and Kannan Ramchandran. 2024.
\newblock Embedllm: Learning compact representations of large language models.
\newblock \emph{arXiv preprint arXiv:2410.02223}.

\end{thebibliography}
